\documentclass[10pt,journal,twocolumn]{IEEEtran}
\usepackage{graphicx}
\usepackage{amstext}
\usepackage{amsmath}
\usepackage{amsfonts,amssymb}
\usepackage{amsmath,tabu}
\usepackage{amssymb}
\usepackage[dvips]{epsfig}
\usepackage{epsfig}
\usepackage[dvips]{graphicx}
\usepackage{multirow}
\usepackage{multirow} 
\usepackage[utf8]{inputenc}
\usepackage[english]{babel}
\usepackage{caption}
\usepackage{subcaption}
\usepackage{float}
\usepackage{ltablex}
\usepackage{algorithmic}
\usepackage{setspace}
\newtheorem{theorem}{Theorem}[section]

\newtheorem{lemma}[theorem]{Lemma}
\def\bw{{\boldsymbol{w}}}

\def\bx{{\boldsymbol{x}}}
\def\br{{\boldsymbol{r}}}
\def\bz{{\boldsymbol{z}}}

\def\bp{{\boldsymbol{p}}}
\def\bP{{\boldsymbol{P}}}
\def\bq{{\boldsymbol{q}}}

\def\bv{{\boldsymbol{v}}}
\def\bh{{\boldsymbol{H}}}

\def\by{{\boldsymbol{y}}}
\def\one{{\boldsymbol{1}}}

\def\bB{{\boldsymbol{B}}}

\def\bI{{\boldsymbol{I}}}

\def\ba{{\boldsymbol{a}}}
\providecommand{\keywords}[1]{\textbf{\textit{Index terms---}} #1}
\begin{document}
\title{PIANO: A Fast Parallel Iterative Algorithm for Multinomial and Sparse Multinomial Logistic Regression}
\author{R.~Jyothi and P.~Babu}
\maketitle
\begin{abstract}
Multinomial Logistic Regression is a well-studied tool for classification and has been widely used in fields like image processing, computer vision and, bioinformatics, to name a few. Under a supervised classification scenario, a Multinomial Logistic Regression model learns a weight vector to differentiate between any two classes by optimizing over the likelihood objective. With the advent of big data, the inundation of data has resulted in large dimensional weight vector and has also given rise to a huge number of classes, which makes the classical methods applicable for model estimation not computationally viable. To handle this issue, we here propose a parallel iterative algorithm: \textbf{P}arallel \textbf{I}terative \textbf{A}lgorithm for Multi\textbf{N}omial L\textbf{O}gistic Regression (\textbf{PIANO}) which is based on the Majorization Minimization procedure, and can parallely update each element of the weight vectors. Further, we also show that \textbf{PIANO} can be easily extended to solve the Sparse Multinomial Logistic Regression problem - an extensively studied problem because of its attractive feature selection property. In particular, we work out the extension of \textbf{PIANO} to solve the Sparse Multinomial Logistic Regression problem with $\ell_{1}$ and $\ell_{0}$ regularizations. We also prove that \textbf{PIANO} converges to a stationary point of the Multinomial and the Sparse Multinomial Logistic Regression problems. Simulations were conducted to compare \textbf{PIANO} with the existing methods, and it was found that the proposed algorithm performs better than the existing methods in terms of speed of convergence.
\end{abstract}
\section{Introduction}
In the field of machine learning and data mining, one of the central goals is to classify the extracted features from the data into different categories or classes using a statistical model \cite{bishop}. That is, given $(\bx_{i},\by_{i})_{1 \leq i\leq n}$, where $\bx_{i} \in \mathbb{R}^{d}$ is a feature vector and $\by_{i} \in \mathbb{R}^{m}$ is  a class label which is represented as a ``1-of-m'' encoding vector i.e., its $i^{th}$ element is equal to one if $\bx$ belongs to class $i$ and is zero otherwise, the task here is to train a statistical model which can predict $\by$ based on $\bx$. A statistical model classifies the feature vector by dividing the input space into decision regions whose boundaries are the decision surfaces. Under a supervised classification scenario, the task of the statistical model is to learn the parameters of these decision surfaces from the training data by using either a probabilistic or deterministic approach (\cite{bishop, book}). Classifiers such as Support Vector Machine \cite{SVM} and Neural Network \cite{nn} take a deterministic approach towards classification wherein they construct a discriminant function which directly assigns each input feature vector $\bx$ to one of the classes. Classifiers based on probabilistic approach, models the posterior probability $P(\by|\bx)$ as a function of parameter of the decision surfaces. There are two approaches to model the posterior probability - the first approach used by the generative classifiers such as Hidden Markov Model \cite{hmm} and naive Bayes classifier \cite{bayes}, learn a model for joint probability $P(\bx,\by)$ and then uses Bayes rule to calculate the posterior probability $P(\by|\bx)$, and the second approach used by discriminative classifiers directly models the posterior probability  - which reduces the number of parameters to be estimated \cite{bishop}. In this paper, we consider the parameter estimation problem of one such discriminative classifier -  Multinomial Logistic Regression (MLR) which has been widely applied in diverse fields such as hyperspectral image classification (\cite{hyperspectral, semisupervised}), text categorization (\cite{text1, text2}) and in biomedical data analysis (\cite{biomedical1, biomedical2}).  \\
MLR is a linear model, i.e., its decision boundaries are defined via linear functions of the feature vector $\bx$ and are represented by hyperplanes. Denoting the parameter of the $i^{th}$ hyperplane as $\bw_{i} \in \mathbf{R}^{{d}\times 1}$, MLR estimates $\{\bw_{i}\}_{i=1}^{m}$ from the training data and then uses it to predict the class labels of the actual data sample. MLR models the posterior probability as the normalized exponential or the softmax transformation of a linear function of the feature vector $\bx$. Then, representing each class label as a Bernoulli random variable, the parameters $\{\bw_{i}\}_{i=1}^{m}$ are estimated by maximizing the log-likelihood function or by minimizing the negative of the log-likelihood function. This minimization problem, as will be shown in the next section, is convex and differentiable but does not have a closed-form solution. Hence, iterative methods are usually employed to estimate the parameters of the MLR classifier. With the advent of big data, it is desirable that these iterative methods are able to cope up with the large dimensions of the feature vector and the huge number of classes. Large scale data is not atypical (\cite{data, large_data_1, large_data_2}) - for example, in the recently released classification data set from \cite{large_data_1}, which is a library containing large scale text classification data, the total number of classes and features were about $12,294$ and $347,256$, respectively. Therefore, the total number of parameters to be learned were about $4,269,165,264$. In such a case, training MLR classifier with an iterative algorithm which sequentially updates the parameters can be time consuming even for a single iteration. \\
Another common issue which occurs during the training of MLR is the problem of over-fitting; wherein the classifier works perfectly well on the training data but works poorly on the test data - this usually occurs when the number of features $d$ is more than the number of training samples $n$ \cite{vapnik}. To prevent over-fit, a standard approach is to regularize - wherein an extra term is added to the log-likelihood function to penalize the weights taking large values. A common penalty term used is the $\ell_{1}$ norm regularizer which not only penalizes the weights taking large values but also promotes sparsity. However, the addition of $\ell_{1}$ norm regularizer makes the log-likelihood minimization problem non-differentiable and hence it is more challenging to solve when compared to the log-likelihood minimization problem of the unregularized MLR classifier.  \\
A conventional algorithm used to estimate the parameters of the unregularized MLR classifier is the Iterative Reweighted Least Squares algorithm \cite{IRLS}. It is based on Newton-Raphson method and involves computing the inverse of a square matrix of size $dm$ at every iteration. This makes the algorithm computationally expensive for large dimension of the feature vector and for large number of classes. Moreover, since this method is based on Newton-Raphson, IRLS requires the objective function to be differentiable and hence cannot be used to estimate the parameters of the regularized MLR classifier. To avoid taking the inverse at every iteration, the authors in \cite{mm} proposed an algorithm based on Majorization Minimization (MM) principle (which will be explained in section.\ref{sec:2}) which as shown in \cite{mm}, can be easily extended to estimate the parameters of the regularized MLR classifier. However, this algorithm sequentially updates the parameters of the MLR classifier. LC algorithm developed by Gopal et.al. \cite{mm_parallel}, also based on the MM principle, semi-parallely updates the parameters of the MLR classifier i.e., it parallely updates the block of weights $\bw_{i}$ corresponding to each class, however similar to the IRLS algorithm, LC requires the objective function to be differentiable and hence cannot be extended to estimate the parameters of the regularized MLR classifier. Also, at every iteration, LC algorithm uses the LBFGS solver, which as shown in the simulation section, hampers its convergence speed. The algorithms proposed in (\cite{admm, scutari}) were developed explicitly to estimate the parameters of the regularized MLR classifier. However, these algorithms are sequential in nature. In this paper, we present a novel fast parallel algorithm (\textbf{PIANO}), which unlike the LC algorithm, updates every element of each $\bw_{i}$ parallely. We also extend the algorithm to estimate the parameters of the regularized MLR classifier. The major contributions of the paper are as follows: 
\begin{enumerate}
\item{A MM based parallel algorithm - \textbf{P}arallel \textbf{I}terative \textbf{A}lgorithm for Multi\textbf{N}omial L\textbf{O}gistic Regression (\textbf{PIANO}) is proposed to estimate the parameters of the MLR classifier. The proposed algorithm updates each element of $\{\bw_{i}\}_{i=1}^{m}$ parallely - which is useful when the number of features and classes are huge. }
\item{We also show that the proposed algorithm can be extended to estimate the parameters of the regularized MLR classifier. We estimate the parameters for both $\ell_{1}$ and $\ell_{0}$ regularizations.}
\item{The monotonicity and convergence to a stationary point is proved for the proposed algorithm.}
\item{Numerical simulations are conducted to compare the proposed algorithms with the existing algorithms.}
\end{enumerate}   
The paper is organized as follows. We formulate the problem and give a brief review of the existing methods in Sec. \ref{sec:1}. Next, we given an overview of MM in Sec. \ref{sec:2}. In Sec. \ref{sec:3}, we propose a parallel algorithm \textbf{PIANO} to solve the problem in (\ref{multinomial_logistic_1}) and also show that \textbf{PIANO} can be extended to solve the Sparse MLR problem in (\ref{sparse_l0}) and (\ref{sparse_l1}). Next, we show that the proposed algorithm converges to the stationary point of the MLR and Sparse MLR problem. In Sec. \ref{sec:5} we compare the algorithms with the existing algorithms via computer simulations and conclude the paper in Sec. \ref{sec:6}. 
\section{Problem formulation and literature survey}\label{sec:1}
Given the training samples $(\bx_{i},\by_{i})_{1 \leq i\leq n}$, MLR models the posterior probability \\
$P\left(y_{i}=1|\bx,\bw_{1},\bw_{2},\cdots,\bw_{m}\right)$ as the softmax transformation of a linear function of the feature vector $\bx$:  \\
\begin{equation}\label{eq:11}
\begin{array}{ll}
P\left(y_{i}=1|\bx,\bw_{1},\bw_{2},\cdots,\bw_{m}\right)\overset{\Delta} = \sigma_{i}\left(\bw_{1},\bw_{2},\cdots,\bw_{m},\bx\right)\\ \hspace{4.5cm} \overset{\Delta} = \dfrac{\textrm{exp}\left(\bw_{i}^{T}\bx\right)}{\displaystyle\sum_{j=1}^{m}\textrm{exp}\left(\bw_{j}^{T}\bx\right)}
\end{array}
\end{equation}
where $y_{i}$ is the $i^{th}$ element of $\by$, $m$ is the number of classes, $\sigma_{i}(\ba) \overset{\Delta} = \dfrac{\textrm{exp}(a_{i})}{\displaystyle\sum_{j=1}^{m}\textrm{exp}(a_{j})}$ is the softmax function.  When $m=2$, the model in (\ref{eq:11}) corresponds to logistic regression model and for $m>2$, the above model is known by several names such as multinomial logistic regression model, softmax regression model, and the conditional maximum entropy model \cite{bishop}. \\
Assuming that the $n$ training samples are generated independently, the components of $\{{\bw_{i}}\}{_{i=1}^{i=m}}$ are learned from the training data $(\bx_{i},\by_{i})_{1 \leq i\leq n}$ using the maximum likelihood approach. Since each class label $\by_{i}$ is a binary vector, they can be modeled as a Bernoulli multivariate random variable \cite{prob}. The likelihood function is given by: 
\begin{equation}\label{likelihood}
\begin{array}{ll}
L({\tilde{\bw}}) =\displaystyle\prod_{j=1}^{n}\displaystyle\prod_{i=1}^{m}\left(P\left(y_{i}=1|\bx_{j},\bw_{1},\bw_{2},\cdots,\bw_{m}\right)\right)^{y_{ji}} \\ \hspace{1cm}=  \displaystyle\prod_{j=1}^{n}\displaystyle\prod_{i=1}^{m}\sigma_{i}\left(\bw_{1},\bw_{2},\cdots,\bw_{m},\bx_{j}\right)^{y_{ji}}
\end{array} 
\end{equation}
where $\tilde{\bw}\in \mathbf{R}^{dm \times 1}$ is obtained by stacking $[\bw_{1}^{T}, \cdots, \bw_{m}^{T}]^{T}$, $y_{ji}$ is used to denote the $i^{th}$ component of the $j^{th}$ class label $\by_{j}$ and $\sigma_{i}\left(\bw_{1},\bw_{2},\cdots,\bw_{m},\bx_{j}\right)$ is given by (\ref{eq:11}). The components of $\{{\bw_{i}}\}{_{i=1}^{i=m}}$ can be estimated by maximizing the log-likelihood or by minimizing the negative of the log-likelihood function in (\ref{likelihood}): 
\begin{equation}\label{multinomial_logistic_1}
\begin{array}{ll}
\textrm{MLR:} \quad \underset{{\tilde{\bw}}}{\rm minimize}\: l_{\textrm{MLR}}({{\tilde{\bw}}})\overset{\Delta} = \\\: \displaystyle\sum_{j=1}^{n}\left(-\sum_{i=1}^{m}y_{ji}\bw_{i}^{T}\bx_{j}+\textrm{log}\sum_{i=1}^{m}\textrm{exp}\left(\bw_{i}^{T}\bx_{j}\right) \right)
\end{array}
\end{equation}
All though the problem in (\ref{multinomial_logistic_1}) is convex and differentiable, one cannot obtain a closed-form solution using the KKT conditions primarily due to the presence of log-sum-exponential terms in (\ref{multinomial_logistic_1}). Hence, iterative methods are usually employed to solve the problem in (\ref{multinomial_logistic_1}). A conventional algorithm used to solve the problem in (\ref{multinomial_logistic_1}) is the Iterative Reweighted Least Squares (IRLS) algorithm \cite{IRLS} which is based on Newton-Raphson method. Its update equation is given by: 
\begin{equation}\label{IRLS}
\begin{array}{ll}
{\tilde{\bw}}^{k+1} ={\tilde{\bw}}^{k} - {\bh\left(\tilde{\bw}^{k}\right)}^{-1}\br\left(\tilde{\bw}^{k}\right)
\end{array}
\end{equation}
where $\tilde{\bw}^{k}$ is the value taken by $\tilde{\bw}$ at the $k^{th}$ iteration, $\br\left(\tilde{\bw}^{k}\right)$ and $\bh\left(\tilde{\bw}^{k}\right)$ are the gradient and Hessian of $l_{\textrm{MLR}}\left(\tilde{\bw}\right)$ at $\tilde{\bw} = \tilde{\bw}^{k}$, respectively and it is given by:  
\begin{equation}\label{gradient}
\begin{array}{ll}
\br\left(\tilde{\bw}^{k}\right)= -\displaystyle\sum_{j=1}^{n}\left(\bp_{j}\left({\tilde{\bw}}^{k}\right)-\by_{j}\right) \otimes \bx_{j}
\end{array}
\end{equation}
\begin{equation}\label{hessian}
\begin{array}{ll}
\bh\left(\tilde{\bw}^{k}\right)= \displaystyle\sum_{j=1}^{n} \left(\bP_{j}\left(\tilde{\bw}^{k}\right)-\bp_{j}\left(\tilde{\bw}^{k}\right)\bp_{j}\left(\tilde{\bw}^{k}\right)^{T}\right)\otimes \bx_{j}\bx_{j}^{T}
\end{array}
\end{equation}
where $\otimes$ represents the Kronecker operator, $\bp_{j}(\tilde{\bw}^{k}) = [p_{j}^{(1)}(\tilde{\bw}^{k}), \cdots p_{j}^{(m)}(\tilde{\bw}^{k})]^{T}$, $p_{j}^{(i)}(\tilde{\bw}^{k}) = P\left(y_{ij}=1|\bx_{j},\bw^{k}_{1},\bw^{k}_{2},\cdots,\bw^{k}_{m}\right)$ and $\bP_{j}\left(\tilde{\bw}^{k}\right)$ is a diagonal matrix with diagonal elements $\{p_{j}^{(1)}(\tilde{\bw}^{k}), \cdots, p_{j}^{(m)}(\tilde{\bw}^{k})\}$. Since the Hessian is a function of $\tilde{\bw}^{k}$, its inverse has to be computed at every iteration - which makes the IRLS algorithm computationally expensive. To avoid taking the inverse at every iteration, the authors in \cite{mm} proposed an algorithm based on Majorization Minimization (MM) principle  with the following update step:\\
\begin{equation}\label{MM}
\begin{array}{ll}
\tilde{\bw}^{k+1} = \tilde{\bw}^{k} - \bB^{-1}\br\left(\tilde{\bw}^{k}\right)
\end{array}
\end{equation}  
where $\bB = \dfrac{1}{2}\left(\bI - \dfrac{\one\one^{T}}{m}\right)\otimes\displaystyle\sum_{j=1}^{n}\bx_{j}\bx_{j}^{T}$ and $\one = (1,1,\cdots,1)^{T}$. The update step in (\ref{MM}) involves computing the inverse of $\bB$ however, since $\bB$ is independent of $\tilde{\bw}$, its inverse can be precomputed - giving it a computational benefit over IRLS. Recently, the authors in \cite{mm_parallel} proposed a semi-parallel algorithm named LC which is also based on MM procedure wherein the weights corresponding to each class can be updated parallely i.e., each $\{{\bw_{i}}\}_{i=1}^{i=m}$ can be updated parallely. At every iteration, the authors in \cite{mm_parallel} solved the following sub-problem: 
\begin{equation}\label{lc_bound}
\begin{array}{ll}
 \underset{\bw_{i}}{\rm arg\, min}\: -\displaystyle\sum_{j=1}^{n}y_{ij}\bw_{i}^{T}\bx_{j}+ \displaystyle\sum_{j=1}^{n}a_{j}\:{\textrm{exp}}\left(\bw_{i}^{T}\bx_{j}\right)
\end{array}
\end{equation}
\begin{equation}
\begin{array}{ll}
{\textrm{where\quad}} a_{j} \overset{\Delta} = \dfrac{1}{\displaystyle\sum_{i=1}^{m}\textrm{exp}\left(\left({\bw_{i}^{k}}\right)^{T}\bx_{j}\right)}
\end{array}
\end{equation}
The problem in (\ref{lc_bound}) does not have a closed-form solution and the authors in \cite{mm_parallel} used the LBFGS solver to obtain $\bw_{i}$, which hampers the speed of the algorithm. The authors in \cite{mm_parallel} have also proposed another parallel algorithm using Alternating Direction Method of Multipliers (ADMM). The numerical simulations in \cite{mm_parallel} report that ADMM has slower speed of convergence when compared to LC. \\
We also consider the problem of parameter estimation of the regularized MLR classifier: 
\begin{equation}\label{sparse_l0}
\begin{array}{ll}
\textrm{S0-MLR:} \quad \underset{\tilde{\bw}}{\rm minimize}\: l_{\textrm{MLR}}{(\tilde{\bw})} + \lambda{\|\tilde{\bw}\|_{0}}
\end{array}
\end{equation}
where $\lambda > 0$ is the regularization parameter and ${\|.\|}_{0}$ is the $\ell_{0}$ vector norm. An attractive feature of the problem in (\ref{sparse_l0}) is that the weights obtained are sparse in nature - which helps in feature selection and also has computational benefits \cite{feature_selection}. The problem in (\ref{sparse_l0}) is non-convex and not differentiable and is usually solved by either approximating the $\ell_{0}$ constraint (\cite{l0_approx_1}, \cite{l0_approx_2}) or by relaxing the $\ell_{0}$ norm with $ \ell_{1}$ norm:
\begin{equation}\label{sparse_l1}
\begin{array}{ll}
\textrm{S1-MLR:} \quad \underset{\tilde{\bw}}{\rm minimize}\: l_{\textrm{MLR}}{(\tilde{\bw})} + \lambda{\|\tilde{\bw}\|_{1}}
\end{array}
\end{equation}
The solution of the above problem can be interpreted as the maximum a posteriori estimate of $\tilde{\bw}$ with the assumption that the elements of $\tilde{\bw}$ has Laplacian prior distribution. Note that one can also use $\ell_{2}$ norm to penalize the large weights. However, it does not result in a sparse weight vector and hence is not usually preferred. The problems in (\ref{sparse_l0}) and (\ref{sparse_l1}) are referred as Sparse MLR problems.   \\
The authors in \cite{mm} have extended their algorithm to solve the problem in (\ref{sparse_l1}). The extended algorithm is based on the combination of MM and alternating minimization i.e., they updated each element of $\tilde{\bw}$ using MM while keeping the other components of $\tilde{\bw}$ fixed. Then they get the following update equation:    \\
\begin{equation}
\begin{array}{ll}
{\tilde{w}_{i}}^{k+1}={\textrm{soft}}\left({\tilde{w}_{i}}^{k}- \dfrac{r_{i}\left(\tilde{\bw}^{k}\right)}{B_{ii}}, \dfrac{\lambda}{B_{ii}}\right)
\end{array}
\end{equation}
where $B_{lm}$ denotes the $(l, m)$ element of matrix $\bB$, $r_{i}(\tilde{\bw}^{k})$ is the $i^{th}$ element of $\br(\tilde{\bw}^{k})$ and 
\begin{equation}
\begin{array}{ll}
{\textrm{soft}}(a, b) = {\textrm{sign}}(a)\, {\textrm{max}}\{0, |a|-b\}
\end{array}
\end{equation}
where $\textrm{max}(a,b)$ chooses the largest value among $a$ and $b$, $\textrm{sign}(a)$ is equal to $1$ if $a >0$ and $\textrm{sign}(a)$ is equal to $-1$ if $a < 0$.
Note that since the above algorithm is based on alternating minimization, one cannot parallely update the elements of $\tilde{\bw}$. The IRLS and the LC algorithm proposed to solve the problem in (\ref{multinomial_logistic_1}) cannot be extended to solve the problem in (\ref{sparse_l1}), as these algorithms requires the objective function to be smooth. Boyd et.al. \cite{admm} proposed a non-parallel algorithm based on ADMM to solve the problem in (\ref{sparse_l1}). They first introduced an additional variable $\tilde{\bz}$ and converted the unconstrained problem in (\ref{sparse_l1}) to a constrained problem:   
\begin{equation}\label{admm}
\begin{array}{ll}
\underset{\tilde{\bw},\:\tilde{\bz}}{\textrm {minimize}}\; l_{\textrm{MLR}}(\tilde{\bw})+g(\tilde{\bz})\\
\textrm{such that} \quad \tilde{\bw}-\tilde{\bz} = \mathbf{0}
\end{array}
\end{equation}
where $g(\tilde{\bz}) = \lambda\|\tilde{\bz}\|_{1}$. Then they formed the augmented lagrangian of the above problem and solved the augmented lagrangian problem by first alternatingly  updating the primal variables $\tilde{\bw}$ and $\tilde{\bz}$. Next, they updated the dual variable using the updated primal variables. The authors in \cite{scutari} proposed GJ-FLEXA, FLEXA and Inexact GJ algorithms to solve the problem in (\ref{sparse_l1}). These algorithms are basically gradient based methods and at every iteration, instead of minimizing the original problem in (\ref{sparse_l1}), these algorithms minimize an approximation of $l_{\textrm{MLR}}{(\tilde{\bw})}$ in (\ref{sparse_l1}). The approximation is done using second-order Taylor series  and an extra term is added to it, to make the approximation strongly convex. The three algorithms differ only in the way they update the elements of $\tilde{\bw}$: GJ-FLEXA is a non-parallel algorithm, while FLEXA and Inexact GJ is a parallel and a hybrid parallel-sequential algorithm, respectively. Under the numerical results section, Facchinei et.al. in \cite{scutari} concluded that the non-parallel algorithm GJ-FLEXA outperforms the other algorithms. This could be because the authors in \cite{scutari} simply approximates the $l_{\textrm{MLR}}(\tilde{\bw})$ using a second-order Taylor series and do not exploit any structure of the objective function in (\ref{sparse_l1}).


\section{Majorization Minimization}\label{sec:2}
\begin{figure}[h]
\centering
\begin{tabular}{c}
\includegraphics[height=2.0in,width=3.4in]{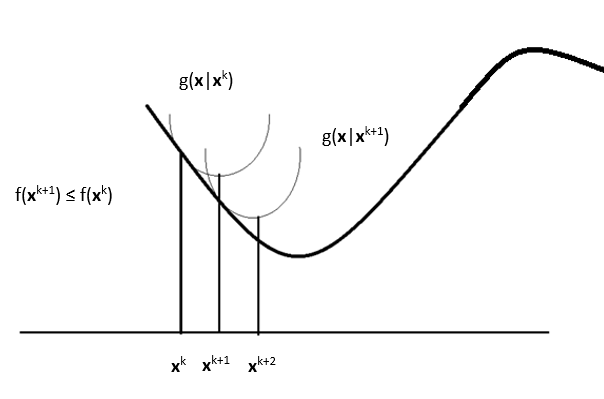}
\end{tabular}
\caption{MM procedure}
\label{mm_procedure}
\end{figure}

Majorization Minimization is a procedure to generate an iterative algorithm which is used to solve an optimization problem $f(\bx)$ more efficiently, for example, in the case of convex problems it can be used to avoid huge matrix inversions and in the case of multivariate optimization problem, it can be used to split the parameters - which allows the algorithm to be implemented parallely. The MM framework mainly consists of two steps: at every iteration construct a ``surrogate'' function $g\left(\bx|\bx^{k}\right)$ which majorizes $f(\bx)$ followed by its minimization to generate $\bx^{k+1}$ i.e.: 
\begin{equation}  \label{eq:mmc}
\bx^{k+1} \in \underset{\bx}{\rm arg\:min} \quad g\left(\bx|\bx^{k}\right)
\end{equation}
The surrogate function is a tighter upper bound of the objective function and hence must satisfy the following properties: 
\begin{equation}  \label{eq:mma}
g\left(\bx^{k}|\bx^{k}\right) = f\left(\bx^{k}\right) 
\end{equation}
\begin{equation}\label{eq:mmb}
g\left(\bx|\bx^{k}\right) \geq f\left(\bx\right) 
\end{equation}
The MM procedure is depicted in Fig. \ref{mm_procedure}, wherein $g(\bx|\bx^{k})$ is the surrogate function which majorizes $f(\bx)$ around $\bx^{k}$ at the $k^{th}$ iteration. From Fig. \ref{mm_procedure}, it can be seen that $f(\bx^{k+2}) < f(\bx^{k+1}) < f(\bx^{k})$. 
By using (\ref{eq:mmc}), (\ref{eq:mma}) and (\ref{eq:mmb}), it can be shown that the objective function is monotonically decreased at every iteration: \\
\begin{equation}\label{eq:24}
f(\bx^{k+1}) \leq g\left(\bx^{k+1}|\bx^{k}\right) \leq g\left(\bx^{k}|\bx^{k}\right) = f(\bx^{k})
\end{equation}
The first inequality and the last equality are by using (\ref{eq:mma}) and (\ref{eq:mmb}). The second inequality is by (\ref{eq:mmc}). Note that an objective function can have more than one surrogate function. However, the computational complexity and convergence rate will vary with the choice of the surrogate function. To have lower computational complexity, the surrogate function must be easy to minimize and the convergence rate of the resultant algorithm will depend on how well the surrogate function follows the shape of the objective function. Hence, the choice of the surrogate function dictates the convergence rate and complexity of the algorithm. An overview of the various surrogate functions can be found in \cite{sir}, \cite{tutorial}.
\section{Proposed Algorithm for Multinomial Logistic Regression}\label{sec:3}
In this section we propose a novel parallel algorithm \textbf{PIANO} to solve the problem in (\ref{multinomial_logistic_1}) based on the MM procedure. The proposed algorithm can parallely update each element of $\tilde{\bw}$ - which is particularly useful when the number of features and classes are large. At the end of this section we prove that the proposed algorithm converges to the stationary point of the problem in (\ref{multinomial_logistic_1}), and discuss its computational complexity. 
\subsection{\textbf{P}arallel \textbf{I}terative \textbf{A}lgorithm for Multi\textbf{N}omial L\textbf{O}gistic Regression (\textbf{PIANO})}\label{mlr}
The objective function $l_{\textrm{MLR}}({{\tilde{\bw}}})$ in (\ref{multinomial_logistic_1}) is not separable in each element of $\tilde{\bw}$ due to the presence of log-sum-exponential terms which couple each element of $\tilde{\bw}$ together - making it challenging to parallely minimize $l_{\textrm{MLR}}({{\tilde{\bw}}})$. In this subsection, we develop an iterative algorithm \textbf{PIANO} which solves the problem in (\ref{multinomial_logistic_1}) using the MM principle in which we form a surrogate function $g(\tilde{w}_{il}|\tilde{\bw}^{k})$ which majorizes the log-sum-exponential term and hence the objective function $l_{\textrm{MLR}}({{\tilde{\bw}}})$ in (\ref{multinomial_logistic_1}). The surrogate function $g(\tilde{w}_{il}|\tilde{\bw}^{k})$ is separable in each element of $\tilde{\bw}$ and hence each element of $\tilde{\bw}$ can be updated parallely. We now discuss the following lemmas which will be used to construct the surrogate function $g(\tilde{w}_{il}|\tilde{\bw}^{k})$. 
\begin{lemma} \label{lemma 1}
Given any ${z= z^{k}}$, $\textrm{log}\left(z\right)$  can be upper  bounded as:
\begin{equation}\label{eq:21}
\begin{array}{ll}
\textrm{log}\left(z\right) \leq  \textrm{log}\left(z^{k}\right) + \dfrac{1}{z^{k}} \left(z-z^{k}\right)
\end{array}
\end{equation}
The upper bound for $\textrm{log}\left(z\right)$ is linear in $z$.
\end{lemma}
\begin{IEEEproof}
Since the log function is concave in $\mathcal{R}$ \cite{boyd}, a tighter upper bound for $\textrm{log}(z)$  at $z^{k}$ can be found by the first order Taylor expansion - which is a tangent plane to the log function at $z=z^{k}$. The first order Taylor approximation for a differentiable function $f(z)$ at $z=z^{k}$ is given by: 
\begin{equation}\label{fote}
\begin{array}{ll}
f(z) \leq  f(z^{k})+ f'\left(z^{k}\right)\left(z-z^{k}\right)
\end{array}
\end{equation}
where $f'\left(z^{k}\right)$ denotes the differentiation of $f(z)$ at $z^{k}$. Substituting for $f(z) = \textrm{log}(z)$ in (\ref{fote}), the inequality in (\ref{eq:21}) is achieved.
\end{IEEEproof}
\begin{lemma} \label{lemma 2}
Given any ${\bw} = {\bw^{k}}$, the function ${\textrm{exp}}\left({\bw^{T}}\bx\right)$ can be upper bounded as:
\begin{equation}\label{eq:21a}
\begin{array}{ll}
{\textrm{exp}}\left({\bw^{T}}\bx\right) \leq \displaystyle\sum_{i=1}^{d}\dfrac{1}{d}\:{\textrm{exp}}\left(d x_{i}\left(w_{i}-w_{i}^{k}\right)+ ({\bw^{k}})^{T}\bx\right)
\end{array}
\end{equation} 
\end{lemma}
\begin{IEEEproof}
We replicate the proof from \cite{sir} for the sake of clarity. Note that the ${\textrm{exp}}\left(\cdot\right)$ is convex and hence by using the Jensen's inequality (\cite{jensen1, jensen2}) we get: 
\begin{equation} \label{jensens}
\begin{array}{ll}
\textrm{exp}\left(\displaystyle\sum_{i=1}^{d} \dfrac{s_{i}}{d}\right) \leq \displaystyle\sum_{i=1}^{d}\dfrac{\textrm{exp}\left(s_{i}\right)}{d}
\end{array}
\end{equation}
Letting $s_{i} = d x_{i}\left(w_{i} - w_{i}^{k}\right) +(\bw^{k})^{T}\bx$ and substituting it in (\ref{jensens}), the inequality in (\ref{eq:21a}) is achieved.
\end{IEEEproof}
Let $z =\displaystyle\sum_{i=1}^{m}\textrm{exp}\left(\bw_{i}^{T}\bx_{j}\right)$, then by using lemma \ref{lemma 1}, we can upper bound the objective in (\ref{multinomial_logistic_1}) at any given $\tilde{\bw}^{k}$ by the following surrogate function $\hat{g}(\bw_{i}|\tilde{\bw}^{k})$:
\begin{equation}\label{upperbound_1}
\begin{array}{ll}
\hat{g}(\bw_{i}|\tilde{\bw}^{k}) =-\displaystyle\sum_{j=1}^{n}\sum_{i=1}^{m}y_{ji}\bw_{i}^{T}\bx_{j}+ \displaystyle\sum_{j=1}^{n}a_{j}\displaystyle \sum_{i=1}^{m} {\textrm{exp}}\left(\bw_{i}^{T}\bx_{j}\right)
\end{array}
\end{equation}
\begin{equation}\label{eq:22}
\begin{array}{ll}
{\textrm{where\quad}}a_{j} = \dfrac{1}{\displaystyle\sum_{i=1}^{m}\textrm{exp}\left(\left({\bw_{i}^{k}}\right)^{T}\bx_{j}\right)}, \: j \in{(1,2\cdots n)}
\end{array}
\end{equation}
\\
Its worth mentioning that the above surrogate function is separable in each $\bw_{i}$. To make it separable in each element of $\bw_{i}$, we once again majorize $\hat{g}(\bw_{i}|\tilde{\bw}^{k})$. Using lemma \ref{lemma 2}, the second term of (\ref{upperbound_1}) can be majorized, after rearranging we arrive at a new upperbound for the objective function in (\ref{multinomial_logistic_1}), which we denote as ${g}\left(w_{il}|\tilde{\bw}^{k}\right)$: 
\begin{equation}\label{eq:28}
\begin{array}{ll}
{g}\left(w_{il}|\tilde{\bw}^{k}\right)= -\displaystyle\sum_{i=1}^{m}\displaystyle\sum_{l=1}^{d}w_{il}v_{il} 
\\+\displaystyle\sum_{j=1}^{n}a_{j} \displaystyle \sum_{i=1}^{m} \sum_{l=1}^{d} \dfrac{1}{d} \left(\dfrac{{\textrm{exp}}\left({d x_{jl}} w_{il}\right)}{{\textrm{exp}}\left(d x_{jl}w_{il}^{k} -\bx_{j}^{T}{\bw_{i}^{k}}\right)}\right)
\end{array}
\end{equation}
where $w_{il}$ denotes the $l^{th}$ component of the $i^{th}$ weight vector $\bw_{i}$ and $v_{il}$ denotes the $l^{th}$ component of $i^{th}$ vector $\bv_{i}$ which is given as:

\begin{equation}\label{eq:27}
\begin{array}{ll}
\bv_{i} =\displaystyle\sum_{j=1}^{n}y_{ji}\bx_{j}, i \in{(1,2\cdots m)}
\end{array}
\end{equation}
Note that $\{{\bv_{i}}\}_{i=1}^{i=m}$ does not depend on the weight matrix and hence can be pre-computed. Also, the surrogate function ${g}\left(w_{il}|\tilde{\bw}^{k}\right)$ is separable in each element of $\tilde{\bw}$. Hence, each $w_{il}$ of $\tilde{\bw}$ can be updated parallely. Therefore, at any iteration, given $\tilde{\bw}^{k}$, the surrogate minimization problem would be:
\begin{equation}\label{sp1}
\begin{array}{ll}
\underset{w_{il}}{\rm minimize} \:-\displaystyle\sum_{i=1}^{m}\displaystyle\sum_{l=1}^{d}w_{il}v_{il} \\+\displaystyle\sum_{j=1}^{n}a_{j} \displaystyle \sum_{i=1}^{m} \sum_{l=1}^{d} \dfrac{1}{d} \left(\dfrac{{\textrm{exp}}\left({d x_{jl}} w_{il}\right)}{{\textrm{exp}}\left(d x_{jl}w_{il}^{k} -\bx_{j}^{T}{\bw_{i}^{k}}\right)}\right)
\end{array}
\end{equation}
The above problem does not have a closed-form solution. Below we propose a parameter free bisection method to solve the problem in (\ref{sp1}). To discuss the same in a clear way, we consider the generic form of the problem in (\ref{sp1}): 
\begin{equation}\label{generic}
\begin{array}{ll}
\underset{w}{\rm minimize} f(w) \overset{\Delta} = \: - wv+\displaystyle\sum_{j=1}^{n} {{r}}_{j} \textrm{exp}\left(x_{j}w\right)
\end{array}
\end{equation}
The gradient of the objective function in (\ref{generic}) is given by:
\begin{equation}\label{eq:29}
\begin{array}{ll}
f'(w) \overset{\Delta} = \: -v + \displaystyle\sum_{j=1}^{n}{{r}}_{j}x_{j} \textrm{exp}\left(x_{j}w\right)
\end{array}
\end{equation}
The gradient in (\ref{eq:29}) can be shown to be always increasing. We exploit this fact to choose the initial interval $[a,b]$ of the bisection method - which otherwise becomes a burden and has to be correctly chosen for the bisection method to work. We now discuss different possibilities to choose $a$ and $b$ based on the value of the gradient at $w= 0$:  
\begin{itemize}
\item{Case 1: $f'(0) > 0$  \\}
Consider the following example: $f_{1}(w) = 10w +\textrm{exp}(5w) + \textrm{exp}(-4w)$ whose gradient is  $f_{1}'(w) = 10 +5\textrm{exp}(5w) - 4\textrm{exp}(-4w)$ which is plotted in Fig. \ref{grad_pos}. 
 \begin{figure}[h]
\centering
\begin{tabular}{c}
\includegraphics[height=2.0in,width=3.5in]{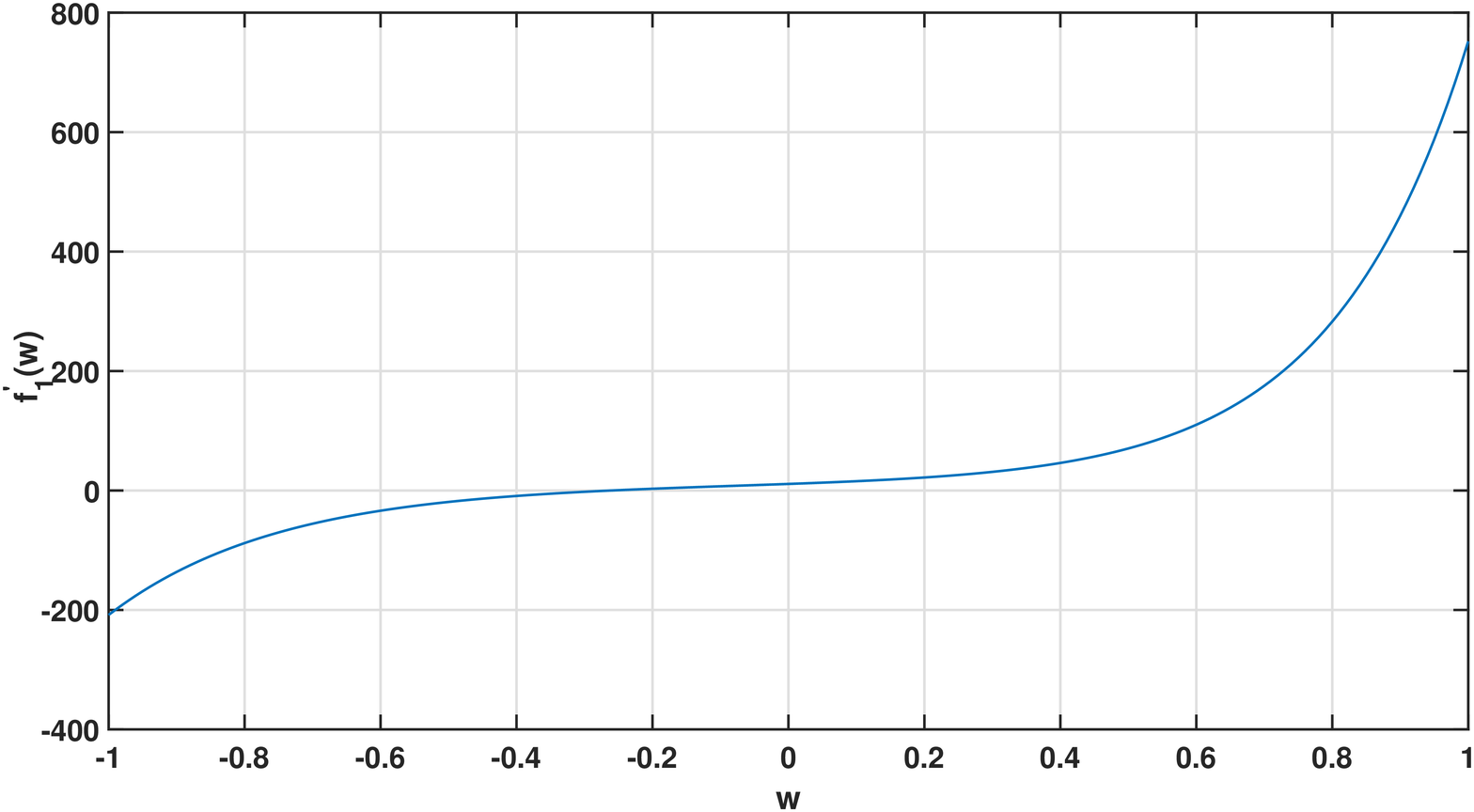}
\end{tabular}
\caption{$f_{1}'(w)$ vs $w$}
\label{grad_pos}
\end{figure}
The following observations can be made from Fig. \ref{grad_pos} - the gradient $f_{1}'(w)$ is always increasing, has a value greater than zero at $w=0$ and is equal to zero for $w=-0.2609$, which is the minimizer. The latter observation is expected because since the gradient is increasing and $f_{1}'(0)$ is positive, the value of the gradient can become equal to zero only for a negative value of $w$. Hence, if one implements bisection method to solve for $f_{1}'(w) = 0$, $a$ can be chosen equal to zero and $b$ could be chosen as a small negative number and one can keep decreasing the value of $b$ until the sign of $f_{1}'(0)$  and sign of $f_{1}'(b)$ are opposite of each other. This makes the choice of $a$ and $b$ simple. 
\item{Case 2: $f'(0) < 0$\\}
Similar to the previous case, to justify the choice of $a$ and $b$ in this case, we start by considering the following example: $f_{2}(w) = -20+\textrm{exp}(3w) + \textrm{exp}(4w)$ whose gradient is  $f_{2}'(w) = -20 +3\textrm{exp}(3w) + 4\textrm{exp}(4w)$ which is plotted in Fig. \ref{grad_neg}. 
 \begin{figure}[h]
\centering
\begin{tabular}{c}
\includegraphics[height=2.0in,width=3.5in]{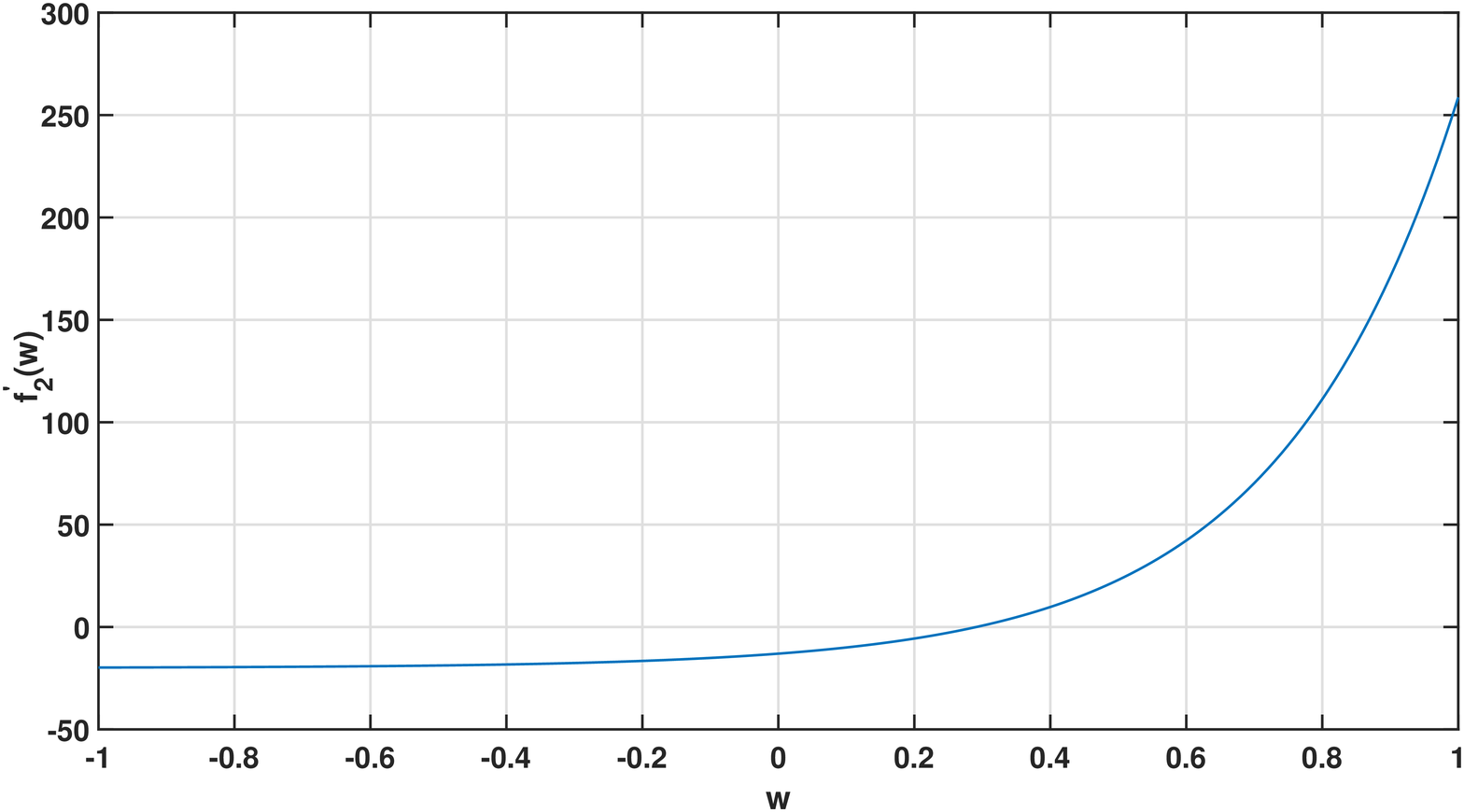}
\end{tabular}
\caption{$f_{2}'(w)$ vs $w$ }
\label{grad_neg}
\end{figure}
Similar to the previous observation, we can see from Fig. \ref{grad_neg} that the gradient of $f_{2}(w)$ is always increasing. However, in contrast to the previous case the gradient has a value lesser than zero at $w=0$ and is equal to zero for $w=0.2911$. This observation is also expected because the gradient is increasing and has a value lesser than zero at $w=0$ - implying that it can become equal to zero only for a positive value of $w$. Hence, one can choose $a=0$ and a small positive number for $b$ and keep increasing it until the sign of $f_{1}'(0)$  and sign of $f_{1}'(b)$ are opposite of each other to solve for $f_{2}'(w) = 0$ using bisection method. 
\end{itemize}
The pseudo code of the proposed algorithm is as follows:
\begin{center}
\begin{tabular}{ p{8.5cm} }
\hline
\hline
\bf{Table 1: Parallel Algorithm for Multinomial Logistic Regression - \textbf{PIANO}} \\
\hline
\hline
{\bf{Input}}: Training samples: ${\bx_{1},\bx_{2}\cdots\bx_{n}}$, Class labels: ${\by_{1},\by_{2}\cdots\by_{n}}$  \\
{\bf{Initialize}}: Set $k=0$. Initialize ${\tilde{\bw}^{0}}$  \\
{\bf{Pre-compute}}: $\bv_{i} =\displaystyle\sum_{j=1}^{n}y_{ij}\bx_{j}, i \in{(1,2\cdots m)}$.\\                 
{\bf{Repeat}}: \\
1) Compute:  $a_{j} = \dfrac{1}{\displaystyle\sum_{i=1}^{m}\textrm{exp}\left(\left({\bw_{i}^{k}}\right)^{T}\bx_{j}\right)},\: j \in{(1,2\cdots n)}$\\
2) Compute the following parallely over all the elements of $\tilde{\bw}$: \\

\, $g'(w_{il}|\tilde{w}^{k})= -v_{il} +\displaystyle\sum_{j=1}^{n}a_{j}x_{jl} \dfrac{\textrm{exp}\left({d x_{jl}} w_{il}\right)}{{\textrm{exp}}\left({d x_{jl}}w_{il}^{k} - \bx_{j}^{T}{\bw_{i}^{k}}\right)} $\\
\, Compute the value of $g'(w_{il}|\tilde{w}^{k})$ at $w_{il} = 0$.\\
\, Choose the value of $b$ based on whether the value of $g'(w_{il}|\tilde{w}^{k})$ at $w_{il} = 0$ is greater than  
 or lesser than zero, as described in the Subsection. \ref{mlr} \\
\, $w_{il}^{k+1}$ is obtained by solving (\ref{sp1}) using bisection method with $a = 0$ and updated $b$.\\
3) $k \leftarrow k+1$, \textbf{until} $\left|\dfrac{{l_{\textrm{MLR}}({{\tilde{\bw}^{k}}})}-{l_{\textrm{MLR}}({\tilde{\bw}^{k-1}})}}{{l_{\textrm{MLR}}({\tilde{\bw}^{k-1}})}}\right| \leq 10^{-3}$\\

\hline
\hline
\end{tabular}
\end{center}
We now discuss the computational complexity of \textbf{PIANO} algorithm. The proposed algorithm \textbf{PIANO}, unlike the IRLS algorithm and the algorithm developed by the authors in \cite{mm}, does not involve computing the inverse of any matrix. Also, when compared to LC which updates each $\bw_{i}$ parallely, \textbf{PIANO} updates every element of each $\bw_{i}$ parallely. Further, each $\{{\bv_{i}}\}_{i=1}^{i=m}$ which is required to compute $w_{il}$ can be pre-computed as it is independent of $w_{il}$ and also at every iteration, except for the computation of $(\bw_{i}^{k})^{T}\bx_{j}$, \textbf{PIANO} requires only inexpensive scalar operations. To solve for the surrogate minimization problem, \textbf{PIANO} implements parameter free bisection method, whose complexity depends on the length of the initial interval $[a,b]$. Since, the value of $a$ and $b$ are chosen such that they are close to the minimizer of the surrogate minimization problem, it reduces the length of the initial interval and thereby reduces the complexity of the bisection method. 
\subsection{Sparse Multinomial Logistic Regression with $\ell_{1}$ regularization}\label{l1 sparse}
In this subsection, we extend the \textbf{PIANO} algorithm to solve the Sparse MLR problem with $\ell_{1}$ regularization, which is given by: 
\begin{equation}\label{l1}
\begin{array}{ll}
\textrm{$\ell_{1}$ Sparse-MLR:} \quad \underset{\tilde{\bw}}{\rm minimize}\: l_{\textrm{MLR}}{(\tilde{\bw})} + \lambda{\|\tilde{\bw}\|_{1}}
\end{array}
\end{equation}
Note that the addition of $\ell_{1}$ norm makes the above problem non-smooth. While the second term of the above problem is already separable in each element of $\tilde{\bw}$, to make the first term of the above problem also separable in each element of $\tilde{\bw}$, we majorize $l_{\textrm{MLR}}{(\tilde{\bw})}$ using lemma \ref{lemma 1} and \ref{lemma 2}, similar to the development of \textbf{PIANO} algorithm:
\begin{equation}\label{eq:41}
\begin{array}{ll}
{g_{_{l1}}}\left(w_{il}|\tilde{\bw}^{k}\right) = -\displaystyle\sum_{i=1}^{m}\displaystyle\sum_{l=1}^{d}w_{il}v_{il} \\
+\displaystyle\sum_{j=1}^{n}a_{j} \displaystyle \sum_{i=1}^{m} \sum_{l=1}^{d} \dfrac{1}{d} \left(\dfrac{{\textrm{exp}}\left({d x_{jl}} w_{il}\right)}{{\textrm{exp}}\left(d x_{jl}w_{il}^{k} -\bx_{j}^{T}{\bw_{i}^{k}}\right)}\right) + \lambda\displaystyle\sum_{i=1}^{m}\displaystyle \sum_{l=1}^{d} \left|w_{il}\right|
\end{array}
\end{equation}
The surrogate function ${g_{_{l1}}}\left(\bw_{il}|\tilde{\bw}^{k}\right)$ is separable in each element of $\tilde{\bw}$ and hence each element of $\tilde{\bw}$ can be updated parallely. Therefore, at any iteration, given $\tilde{\bw}=\tilde{\bw}^{k}$, the surrogate minimization problem is:
\begin{equation}\label{sp2}
\begin{array}{ll}
\underset{w_{il}}{\rm minimize} \:-\displaystyle\sum_{i=1}^{m}\displaystyle\sum_{l=1}^{d}w_{il}v_{il} \\+\displaystyle\sum_{j=1}^{n}a_{j} \displaystyle \sum_{i=1}^{m} \sum_{l=1}^{d} \dfrac{1}{d} \left(\dfrac{{\textrm{exp}}\left({d x_{jl}} w_{il}\right)}{{\textrm{exp}}\left(d x_{jl}w_{il}^{k} -\bx_{j}^{T}{\bw_{i}^{k}}\right)}\right) + \lambda\displaystyle\sum_{i=1}^{m}\displaystyle \sum_{l=1}^{d} \left|w_{il}\right|
\end{array}
\end{equation}
The above problem does not have a closed-form solution. Similar to the previous section, we develop parameter free bisection method to solve the above problem which we explain by considering the following generic problem:
\begin{equation}\label{generic_sparse}
\begin{array}{ll}
\underset{w}{\rm minimize} f(w) \overset{\Delta} = \: - wv+\displaystyle\sum_{j=1}^{n} {{r}}_{j} \textrm{exp}\left(x_{j}w\right) + \lambda\left|w\right|
\end{array}
\end{equation} 
The gradient of the objective function in (\ref{generic_sparse}) is given by:
\begin{equation}\label{eq:42}
\begin{array}{ll}
f'(w) \overset{\Delta} =  -v + \displaystyle\sum_{j=1}^{n}{{r}}_{j}x_{j} \textrm{exp}\left(x_{j}w\right) +\lambda \dfrac{\partial{|w|}}{\partial{w}} = h(w) + \lambda \dfrac{\partial{|w|}}{\partial{w}}
\end{array}
\end{equation}
where the subgradient $\dfrac{\partial{|w|}}{\partial{w}}$ is given as: 
\begin{equation} \label{eq:subval}
\begin{array}{ll}
\dfrac{\partial{|w|}}{\partial{w}}=\left\{ \begin{array}{ll}
1 & \textrm{if}\ w>0\\
-1 & \textrm{if}\ w<0\\
\left[-1,1\right] & \textrm{if} \ w=0
\end{array}\right..
\end{array}
\end{equation}
The gradient in (\ref{eq:42}) is the same as in (\ref{eq:29}), except for the addition of subgradient term whose value can be either $1$, $-1$ or some value in the interval $[-1,1]$ and hence the gradient in (\ref{eq:42}), similar to the gradient in (\ref{eq:29}), is always increasing. We exploit this fact to choose the appropriate subgradient value from (\ref{eq:subval}) and also to choose the initial interval of the bisection method. We now discuss some cases based on the value of $h(0)$ in (\ref{eq:42}):
\begin{itemize}
\item{Case 1: The value of $h(0)$ is greater than one \\}
Consider the following example: $f_{3}(w) =  10w +\textrm{exp}(5w) + \textrm{exp}(-4w) + \left|w\right|$ whose gradient is  $f_{3}'(w) = h(w)+ \dfrac{\partial{|w|}}{\partial{w}} = 10 +5\textrm{exp}(5w) - 4\textrm{exp}(-4w) + \dfrac{\partial{|w|}}{\partial{w}}$. In this case, the solution for  $f_{3}'(w) = 0$ cannot be at $w=0$, since $f_{3}'(0) = h(0) + \dfrac{\partial{|0|}}{\partial{w}}= h(0) \pm 1 \neq 0$. Hence, the solution for $f_{3}'(w) = 0$ can only occur at a positive value of $w$ or at a negative value of $w$, which dictates the value of the subgradient $\dfrac{\partial{|w|}}{\partial{w}}$. The function $h(w)$, plotted in Fig. \ref{grad_pos},  is always increasing and  is greater than one at $w = 0$, which implies that the solution of $f_{3}'(w) =0$ can only be at a negative value of $w$ and hence the value of the subgradient $\dfrac{\partial{|w|}}{\partial{w}}=-1$. Then as discussed in the previous section, to solve for $f_{3}'(w)=  h(w)+ \dfrac{\partial{|w|}}{\partial{w}} = 10 +5\textrm{exp}(5w) - 4\textrm{exp}(-4w) -1 = 0$ using bisection method, $a$ can be chosen equal to zero and $b$ could be chosen as a small negative number and one can keep decreasing the value of $b$ until the sign of $f_{3}'(0)$  and sign of $f_{3}'(b)$ are opposite of each other. 
\item{Case 2: The value of $h(0)$ is lesser than $-1$ \\}
To rationalize the choice of the initial interval and the value of the subgradient in this case, we consider the following example: $f_{4}(w) = -20+\textrm{exp}(3w) + \textrm{exp}(4w) + \left|w\right|$ whose gradient is  $f_{4}'(w) =  h(w)+ \dfrac{\partial{|w|}}{\partial{w}} =   -20 +3\textrm{exp}(3w) + 4\textrm{exp}(4w) + \dfrac{\partial{|w|}}{\partial{w}}$. Similar to the previous case, the solution for $f_{4}'(w) = 0$ cannot be at $w=0$, since $f_{4}'(0) = h(0) \pm 1 \neq 0$. Since $h(0)$ is lesser than $-1$ and the function $h(w)$ is always increasing (as plotted in Fig. \ref{grad_neg}), the solution for $f_{4}'(w) = 0$ can only be at a positive value of $w$ which implies that the value of the subgradient $\dfrac{\partial{|w|}}{\partial{w}}=1$. Hence, in this case, to solve for $f_{4}'(w) = 0$ using bisection method, one can choose $a=0$ and a small positive number for $b$ and keep increasing it until the sign of $f_{4}'(0)$  and sign of $f_{4}'(b)$ are opposite of each other. 
\item{Case 3: The value of $h(0)$ is inbetween $-1$ and $1$ i.e $-1\leq h(0) \leq1$ \\}
If $-1\leq h(0) \leq1$, then $w=0$ is the solution of the problem in (\ref{generic_sparse}). To explain the same, consider the following example: $f_{5}(w) = \textrm{exp}(5w) + \textrm{exp}(-4w) + \left|w\right|$ whose gradient is  $f_{5}'(w) =  h(w)+ \dfrac{\partial{|w|}}{\partial{w}} =   5\textrm{exp}(5w) - 4\textrm{exp}(-4w) + \dfrac{\partial{|w|}}{\partial{w}}$. We have plotted the function $h(w)$ in Fig. \ref{grad_sol_zero}. From the figure, it can be seen that $h(0) = 1$ and hence the solution for $f_{5}'(w)=0$ can either be at $w=0$ or for a negative value of $w$ such that $h(w) = 1$. But as can be seen from Fig. \ref{grad_sol_zero}, only for $w=0$, $h(0) =1$ and hence the only possible solution for $f_{5}'(w)=0$ is at $w=0$. 
\begin{figure}[h]
\centering
\begin{tabular}{c}
\includegraphics[height=2.3in,width=3.5in]{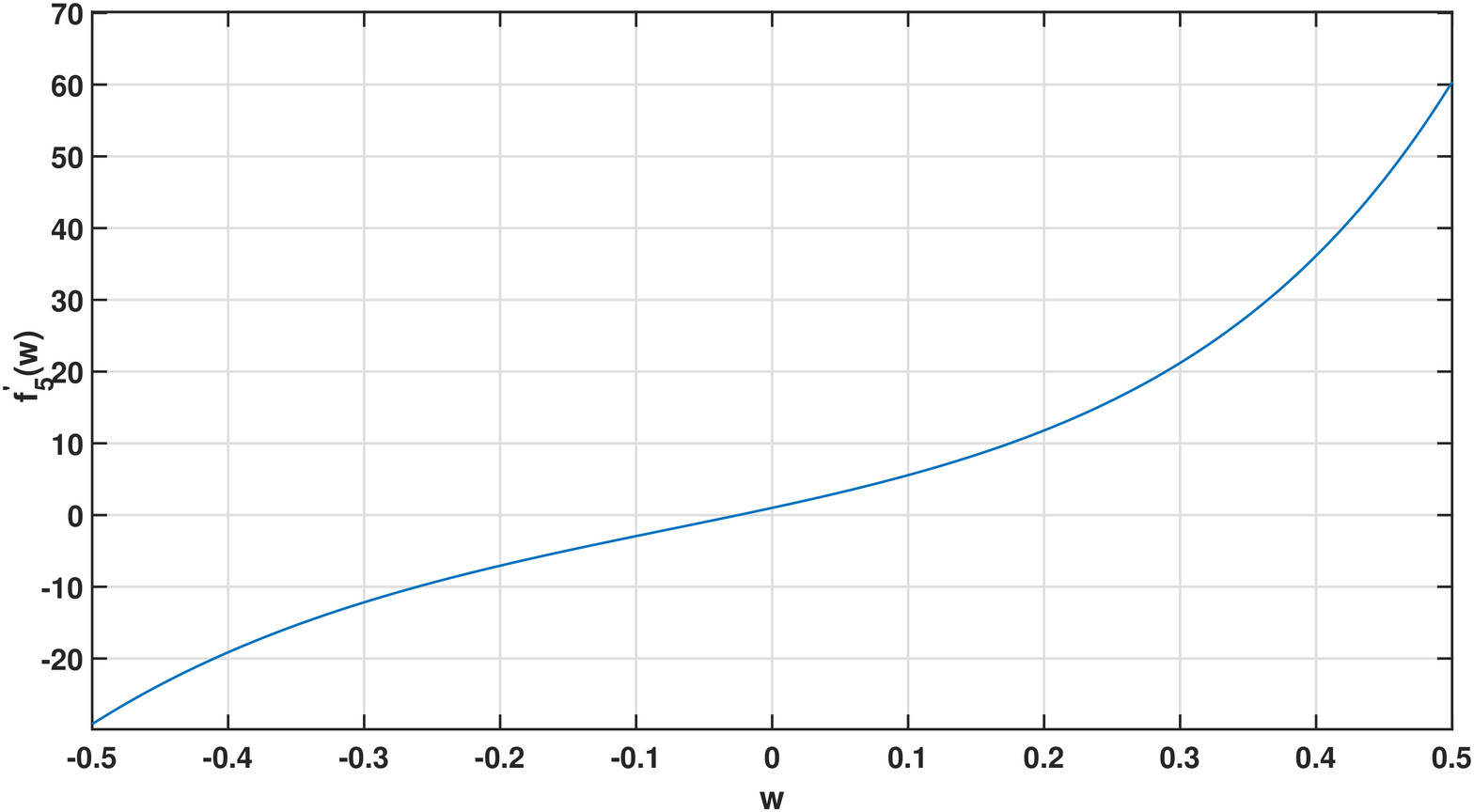}
\end{tabular}
\caption{$f_{5}'(w)$ vs $w$}
\label{grad_sol_zero}
\end{figure}

\end{itemize}
The pseudo code of the proposed algorithm used to solve the sparse MLR problem with $\ell_{1}$ regularization is shown in Table. 2: 
\begin{center}
\begin{tabular}{ p{8.5cm} }
\hline
\hline
\bf{Table 2: Parallel Algorithm for Sparse Multinomial Logistic Regression with $\ell_{1}$ regularization} \\
\hline
\hline
{\bf{Input}}: Training samples: ${\bx_{1},\bx_{2}\cdots\bx_{n}}$, Class labels: ${\by_{1},\by_{2}\cdots\by_{n}}$  \\
{\bf{Initialize}}: Set $k=0$. Initialize ${\tilde{\bw}^{0}}$ and the initial bisection values $a = 0$, $b = 1$  \\
{\bf{Pre-compute}}: $\bv_{i} =\displaystyle\sum_{j=1}^{n}y_{ij}\bx_{j}, i \in{(1,2\cdots m)}$.\\                 
{\bf{Repeat}}: \\
1) Compute:  $a_{j} = \dfrac{1}{\displaystyle\sum_{i=1}^{m}\textrm{exp}\left(\left({\bw_{i}^{k}}\right)^{T}\bx_{j}\right)},\: j \in{(1,2\cdots n)}$\\
2) Compute the following parallely over all the elements of $\tilde{\bw}$: \\
\,\, $h(w_{il})= -\dfrac{v_{il}}{\lambda}+\dfrac{1}{\lambda}\displaystyle\sum_{j=1}^{n}a_{j}x_{jl} \dfrac{\textrm{exp}\left({d x_{jl}} w_{il}\right)}{{\textrm{exp}}\left({d x_{jl}}w_{il}^{k} - \bx_{j}^{T}{\bw_{i}^{k}}\right)} $ \\
%
\hline
\hline
\end{tabular}
\end{center}
\begin{center}
\begin{tabular}{ p{8.5cm} }
\hline
\hline
\bf{Table 2: Parallel Algorithm for Sparse Multinomial Logistic Regression with $\ell_{1}$ regularization} \\
\hline
\hline
\,\, \begin{math} 
\begin{array}{ll}
\dfrac{\partial{|w_{il}|}}{\partial{w_{il}}}=\left\{ \begin{array}{ll}
-1 & \textrm{if}\ h(w_{il}) >1\\
1 & \textrm{if}\  h(w_{il}) <-1\\
\end{array}\right..

\end{array}
    \end{math} 
\\
\,\,\, \textbf{if} $-1\leq h(w_{il})\leq1$ \textbf{then} $w_{il}^{k+1} = 0$ \textbf{else}\\
\,\, $ {g'_{_{l1}}}\left(w_{il}|\tilde{\bw}^{k}\right)= h(w_{il}) + \dfrac{\partial{|w_{il}|}}{\partial{w_{il}}} $\\
\,\, Compute the value of ${g'_{_{l1}}}\left(w_{il}|\tilde{\bw}^{k}\right)$ at $w_{il} = 0$.\\
\,\, Choose the value of $b$ based on whether the value of ${g'_{_{l1}}}\left(w_{il}|\tilde{\bw}^{k}\right)$ at $w_{il} = 0$ is greater than 
\,\, or lesser than zero, as described in Subsection. \ref{mlr}. \\
\,\,\, $w_{il}^{k+1}$ is obtained by solving (\ref{sp2}) using bisection method with $a = 0$ and updated $b$\\
\,\,\, \textbf{end if}\\
3) $k \leftarrow k+1$ \\
\textbf{until} \\
$\left|\dfrac{\left({l_{\textrm{MLR}}({{\tilde{\bw}^{k}}})+ \lambda{\|\tilde{\bw}^{k}\|_{1}}}\right)- \left({l_{\textrm{MLR}}({\tilde{\bw}^{k-1}})}+\lambda{\|\tilde{\bw}^{k-1}\|_{1}}\right)}{\left({l_{\textrm{MLR}}({\tilde{\bw}^{k-1}})}+\lambda{\|\tilde{\bw}^{k-1}\|_{1}}\right)}\right| \leq 10^{-3}$\\
\hline
\hline
\end{tabular}
\end{center}
The \textbf{PIANO} algorithm extended to solve the sparse MLR problem has almost the same complexity as the \textbf{PIANO} algorithm developed to solve the MLR problem with the exception that the former requires an additional computation of $\dfrac{\partial{|w_{il}|}}{\partial{w_{il}}}$, which is computationally inexpensive. Also, when compared to the algorithms developed in \cite{mm} and \cite{admm}, the extended \textbf{PIANO} algorithm can update for each element of $\tilde{\bw}$ parallely.
 
\subsection{Sparse Multinomial Logistic Regression with $\ell_{0}$ regularization}
In this subsection we show that the \textbf{PIANO} algorithm can be extended to solve the sparse multinomial logistic regression problem with $\ell_{0}$ regularization. Like the $\ell_{1}$ regularizer, the $\ell_{0}$ regularizer induces sparsity in $\bw$. The sparse multinomial logistic regression problem with $\ell_{0}$ regularization is given by:
\begin{equation}\label{l0}
\begin{array}{ll}
\textrm{$\ell_{0}$ Sparse-MLR:} \quad \underset{\tilde{\bw}}{\rm minimize}\: l_{\textrm{MLR}}{(\tilde{\bw})}\\
\quad\quad \quad \quad\quad \quad \quad\textrm{subject to} \:\|\tilde{\bw}\|_{0} \leq \beta
\end{array}
\end{equation}
where $\|\tilde{\bw}\|_{0}$ counts the number of non-zero elements in $\tilde{\bw}$ and the constraint is such that $\tilde{\bw}$ must not have more than $\beta$ non-zero elements. The problem in (\ref{l0}) in contrast to the problem in (\ref{l1}) is both non-convex and non-smooth and it is challenging to solve. We now show that the \textbf{PIANO} algorithm can be applied to solve the problem in (\ref{l0}).  We first majorize $l_{\textrm{MLR}}{(\tilde{\bw})}$ as discussed in subsection \ref{mlr} using lemma \ref{lemma 1} and lemma \ref{lemma 2}. Then at any iteration, given $\tilde{\bw}^{k}$, the surrogate minimization problem becomes:
\begin{equation}\label{eq:51}
\begin{array}{ll}
\underset{w_{il}}{\rm minimize}\:-\displaystyle\sum_{i=1}^{m}\displaystyle\sum_{l=1}^{d}w_{il}v_{il}\\ +\displaystyle\sum_{j=1}^{n}a_{j} \displaystyle \sum_{i=1}^{m} \sum_{l=1}^{d} \dfrac{1}{d} \left(\dfrac{{\textrm{exp}}\left({d x_{jl}} w_{il}\right)}{{\textrm{exp}}\left(d x_{jl}w_{il}^{k} -\bx_{j}^{T}{\bw_{i}^{k}}\right)}\right)\\
\\
\hspace{2cm} \textrm{subject to} \:\|\tilde{\bw}\|_{0} \leq \beta
\end{array}
\end{equation}
The problem above does not enjoy a closed-form solution. Note that the above problem without the constraint is the surrogate minimization problem in  (\ref{sp1}) used to solve the MLR problem. Hence, we first minimize ${g}\left(w_{il}|\tilde{\bw}^{k}\right)$ without the sparsity constraint using the bisection approach as discussed in subsection \ref{mlr} and to satisfy the constraint in (\ref{eq:51}) we evaluate ${g}\left(w_{il}|\tilde{\bw}^{k}\right)$ at its minimizer $w^{*}_{il}$ and sort the values in ascending order. Then we preserve the elements of $\tilde{\bw}^{*}$ corresponding to the first $\beta$ values of the sorted ${g}\left(w^{*}_{il}|\tilde{\bw}^{k}\right)$ and assign the remaining elements of $\tilde{\bw}^{*}$ to zero.

\begin{center}
\begin{tabular}{ p{8.5cm} }
\hline
\hline
\bf{Table 3: Parallel Algorithm for Sparse Multinomial Logistic Regression with $\ell_{0}$ regularization} \\
\hline
\hline
{\bf{Input}}: Training samples: ${\bx_{1},\bx_{2}\cdots\bx_{n}}$, Class labels: ${\by_{1},\by_{2}\cdots\by_{n}}$ \\
{\bf{Initialize}}: Set $k=0$. Initialize ${\tilde{\bw}^{0}}$. \\          
{\bf{Repeat}}: \\
1) Compute the following parallely over all the elements of $\tilde{\bw}$ using 
the user-independent bisection\\ method developed in subsection \ref{mlr}:\\
$w^{*}_{il} = \underset{{w_{il}}}{\rm minimize}\:-\displaystyle\sum_{i=1}^{m}\displaystyle\sum_{l=1}^{d}w_{il}v_{il} +\displaystyle\sum_{j=1}^{n}a_{j} \displaystyle \sum_{i=1}^{m} \sum_{l=1}^{d} \dfrac{1}{d} \left(\dfrac{{\textrm{exp}}\left({d x_{jl}} w_{il}\right)}{{\textrm{exp}}\left(d x_{jl}w_{il}^{k} -\bx_{j}^{T}{\bw_{i}^{k}}\right)}\right)$\\
2) Preserve the elements of $\tilde{\bw}^{*}$ corresponding to the first $\beta$ values of the sorted ${g}\left(w^{*}_{il}|\tilde{\bw}^{k}\right)$. \\
3) Assign the remaining elements of $\tilde{\bw}^{*}$ to zero.\\
4) $\tilde{\bw}^{k+1} = \tilde{\bw}^{*}$\\
5) $k \leftarrow k+1$\\
\textbf{until} $\left|\dfrac{\left({l_{\textrm{MLR}}({{\tilde{\bw}^{k}}})+ \beta{\|\tilde{\bw}^{k}\|_{0}}}\right)- \left({l_{\textrm{MLR}}({\tilde{\bw}^{k-1}})}+\beta{\|\tilde{\bw}^{k-1}\|_{0}}\right)}{\left({l_{\textrm{MLR}}({\tilde{\bw}^{k-1}})}+\beta{\|\tilde{\bw}^{k-1}\|_{0}}\right)}\right| \leq 10^{-3}$\\
\hline
\hline
\end{tabular}
\end{center}
Note that the \textbf{PIANO} algorithm extended to solve the sparse MLR problem with $\ell_{0}$ regularization has the computational complexity as \textbf{PIANO} with a small additional complexity due to the sorting step done to satisfy the constraint in (\ref{l0}). 
\subsection{Proof of Convergence for \textbf{PIANO}}
Given that \textbf{PIANO} is based on MM procedure, the sequence of points $\{\tilde{\bw}^{k}\}$ generated by MM algorithm will monotonically decrease the problem in (\ref{multinomial_logistic_1}). Moreover, since $l_{\textrm{MLR}}({\tilde{\bw}})$ in (\ref{multinomial_logistic_1}) is bounded below, it is ensured that the sequence $l_{\textrm{MLR}}({\tilde{\bw}^{k}})$ will converge to a finite value. \\
We now show that the sequence $\{\tilde{\bw}^{k}\}$ converges to the stationary point of the problem in (\ref{multinomial_logistic_1}). Firstly, from the monotonic property of MM we have: 
\begin{equation}\label{eq:conv}
\begin{array}{ll}
l_{\textrm{MLR}}({\tilde{\bw}^{0}})\geq l_{\textrm{MLR}}({\tilde{\bw}^{1}})\geq l_{\textrm{MLR}}({\tilde{\bw}^{2}})
\end{array}
\end{equation}
Assume that there is a subsequence $\tilde{\bw}^{r_{j}}$ converging to a limit point $\tilde{\bq}$. Then from (\ref{eq:mma}), (\ref{eq:mmb}) and  (\ref{eq:conv}) we get:
\begin{equation}
\begin{array}{ll}
g(\tilde{\bw}^{r_{j+1}}|\tilde{\bw}^{r_{j+1}}) = l_{\textrm{MLR}}({\tilde{\bw}^{r_{j+1}}}) \leq l_{\textrm{MLR}}({\tilde{\bw}^{r_{j}+1}}) \leq g(\tilde{\bw}^{r_{j}+1}|\tilde{\bw}^{r_{j}}) \leq g(\tilde{\bw}|\tilde{\bw}^{r_{j}})
\end{array}
\end{equation}
where $g(.)$ is the surrogate function as defined in (\ref{eq:28}). Then, letting $j \rightarrow \infty$, we get:
\begin{equation}
\begin{array}{ll}
g(\tilde{\bq}|\tilde{\bq})  \leq g(\tilde{\bw}|\tilde{\bq})
\end{array}
\end{equation}
which implies $g'(\tilde{\bq}|\tilde{\bq}) \geq 0$. Since the first order behavior of surrogate function is same as function $l_{\textrm{MLR}}({\tilde{\bw}})$ (\cite{convergence}), $g'(\tilde{\bq}|\tilde{\bq}) \geq 0$  implies $l'_{\textrm{MLR}}({\tilde{\bq}}) \geq 0$. Hence, $\tilde{\bq}$ is the stationary point of $l_{\textrm{MLR}}({\tilde{\bw}})$ and therefore the proposed algorithm converges to the stationary point of the problem in (\ref{multinomial_logistic_1}).\\
Similar analysis can be done to show that the proposed algorithm converges to the stationary point of the sparse multinomial logistic regression problem with $\ell_{1}$ and $\ell_{0}$ regularization, since both the problems are bounded below and are solved using MM procedure. Hence, we do not discuss their proof of convergence in detail here. 
\section{Performance study: simulations and real-life data set}\label{sec:5}
\begin{figure}[h]
\centering
\begin{tabular}{c}
\includegraphics[height=2.3in,width=3.5in]{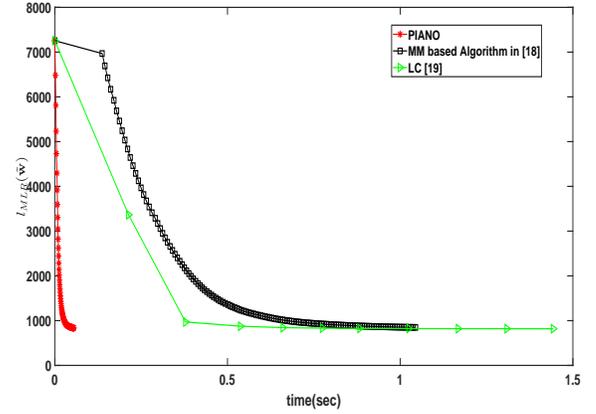}
\end{tabular}
\caption{Comparison of convergence speeds of the proposed algorithm with the MM based algorithm developed in \cite{mm} and the LC algorithm developed in \cite{mm_parallel} .}
\label{monotonic}
\end{figure}

\begin{figure}[!h]
\centering
\begin{tabular}{c}
\includegraphics[height=2.3in,width=3.5in]{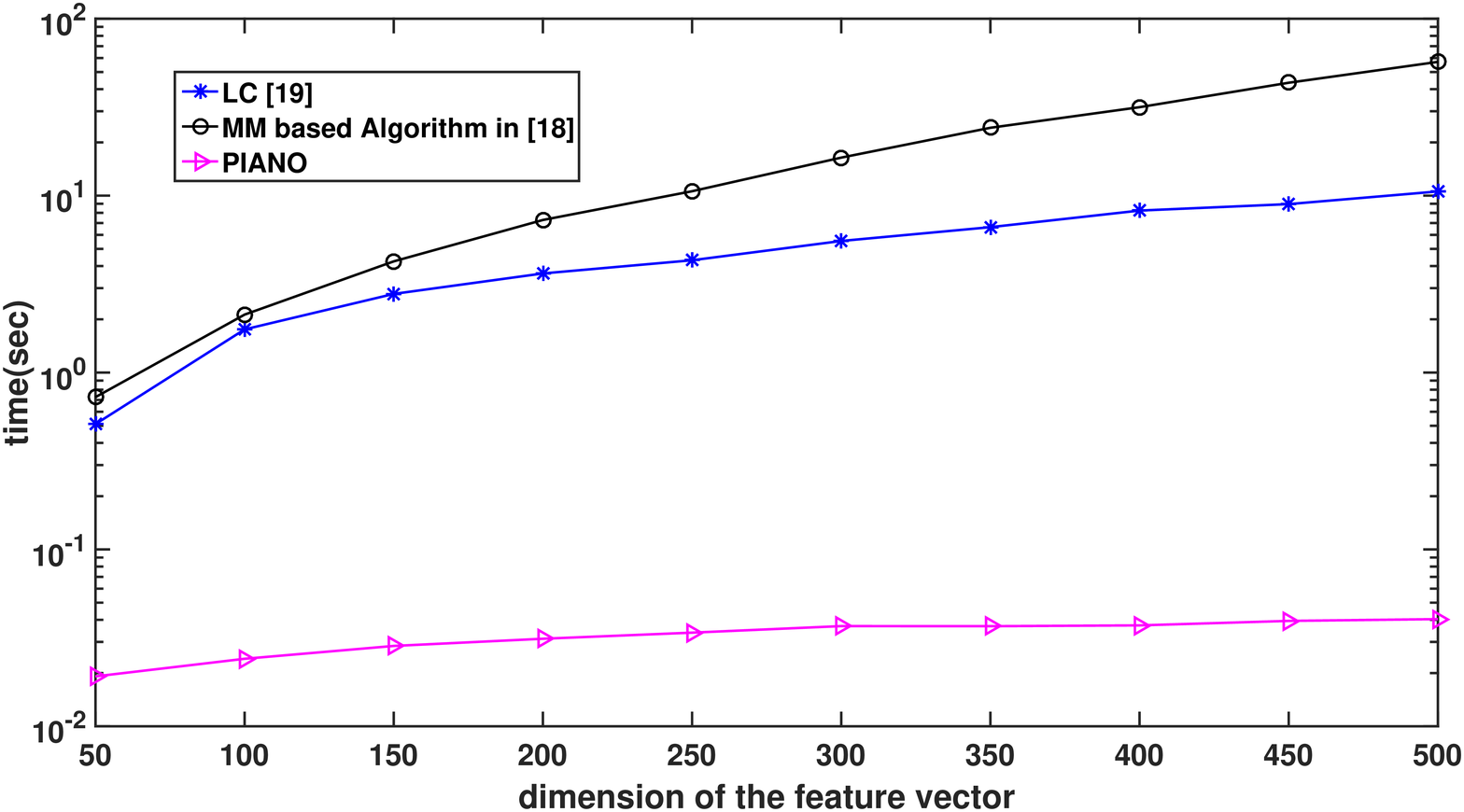}
\end{tabular}
\caption{Comparison of run time of proposed algorithm with existing algorithm for varying dimension $d$ of the feature vector.}
\label{vary_d}
\end{figure}

\begin{figure}[h]
\centering
\begin{tabular}{c}
\includegraphics[height=2.3in,width=3.5in]{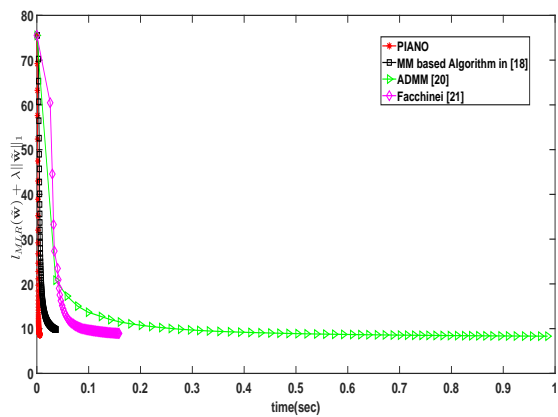}
\end{tabular}
\caption{Comparison of convergence rate of the proposed algorithm with MM based algorithm developed in \cite{mm}, ADMM algorithm proposed in \cite{admm} and the algorithm developed in \cite{scutari} by Facchinei et.al.}
\label{monotonic_sparse}
\end{figure}

\begin{figure}[!h]
\centering
\begin{tabular}{c}
\includegraphics[height=2.3in,width=3.5in]{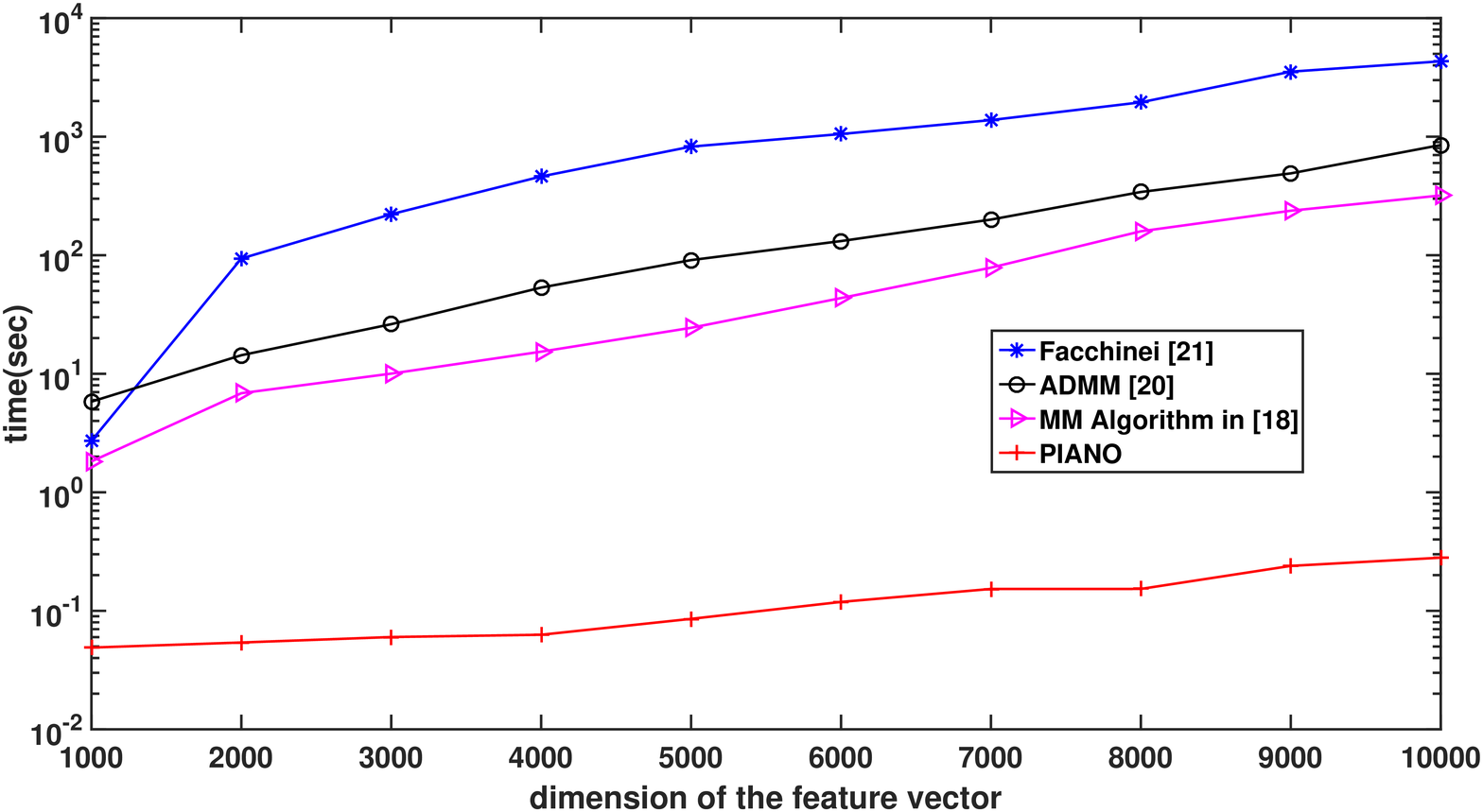}
\end{tabular}
\caption{Comparison of run time of proposed algorithm with existing algorithm for varying dimension $d$ of the feature vector.}
\label{vary_d_sparse}
\end{figure}

\begin{figure}[!h]
\centering
\begin{tabular}{c}
\includegraphics[height=2.3in,width=3.5in]{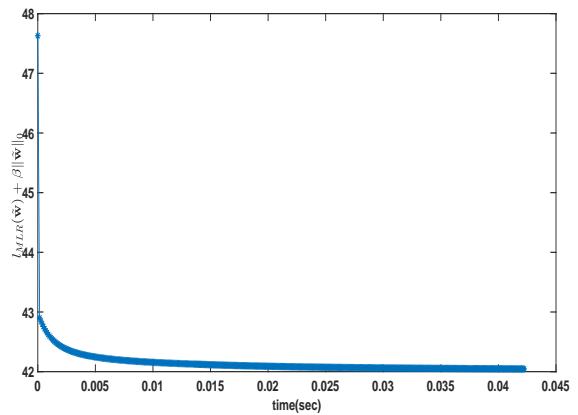}
\end{tabular}
\caption{Monotonic convergence of \textbf{PIANO} for Sparse MLR with $\ell_{0}$ regularization}
\label{l1_l0}
\end{figure}

In this section we present numerical simulations to compare the \textbf{PIANO} algorithm with the state-of-the art algorithms used to solve the MLR and Sparse MLR problems. In particular, for the MLR problem we compare \textbf{PIANO} with the MM based algorithm developed in \cite{mm} and the class wise semi-parallel LC algorithm \cite{mm_parallel}. In the case of Sparse MLR, we compare the proposed algorithm with the MM based algorithm developed in \cite{mm}, ADMM \cite{admm} and the algorithm developed in \cite{scutari}. All the simulations were carried out on a PC with 2.40GHz Intel Xeon Processor with 64 GB RAM.\\
A \:\emph{Multinomial Logistic Regression}\\
a) In the first simulation, we fix the dimension of the feature vector $d$ to be equal to $50$, the number of samples $n$ to be $500$ and the number of classes $m$ to be $30$ and compare the convergence speed of the proposed algorithm with the state-of-the art algorithms, the MM based algorithm proposed in \cite{mm} and the class wise semi-parallel algorithm LC proposed in \cite{mm_parallel}. The elements of $\bx_{j}$ was randomly generated from Standard Normal distribution with zero mean and unit variance. The algorithms were made to run until the following condition was met: 
\begin{equation}\label{condition}
\begin{array}{ll}
\left|\dfrac{{f({\tilde{\bw}^{k}})}-{f({\tilde{\bw}^{k-1}})}}{{f({\tilde{\bw}^{k-1}})}}\right| \leq 10^{-3}
\end{array}
\end{equation}
where $f({\tilde{\bw}^{k}})$ stands for $l_{\textrm{MLR}}({\tilde{\bw}})$. The initial objective value ${l_{\textrm{MLR}}({\tilde{\bw}^{0}})}$ for all the three algorithms were kept same. Fig.\ref{monotonic} shows the run time vs objective value in log for the three algorithms. 
From Fig. \ref{monotonic}, it can be seen that the proposed algorithm takes lesser time to converge when compared to the other two algorithms. Although, it is a single run, we observed the same for different values of problem settings. This is mainly due to the parallel nature of the proposed algorithm which can parallely update each element of $\tilde{\bw}$ when compared to the class wise semi-parallel algorithm - LC and the non-parallel MM algorithm. \\
\\
b) In this simulation, we vary the size of $\tilde{\bw}$ and compare the performance of our algorithm with other standard methods. The comparison is done based on how quickly the algorithms reduce the initial objective value ${l_{\textrm{MLR}}({\tilde{\bw}^{0}})}$ to about $60\%$ of the initial objective value ${l_{\textrm{MLR}}({\tilde{\bw}^{0}})}$. The dimension $d$ was varied from $50$ to $500$ in steps of $50$, the number of samples $n$ and the number of classes $m$ were equal to $1000$ and $30$, respectively. The elements of $\bx_{j}$ was randomly generated from a Standard Normal distribution with zero mean and unit variance. The initial value of $\tilde{\bw}$ was randomly generated from a uniform distribution from $[0,1]$ and was kept same for all the three algorithms. The run time was averaged over $50$ trials. Fig. \ref{vary_d} shows the performance of the algorithms for varying dimension $d$, number of samples $n$ equal to $1000$ and number of classes $m$ equal to $30$. 
From Fig. \ref{vary_d} it can be seen that the proposed algorithm takes the least time to converge to $60\%$ of the initial objective value ${l_{\textrm{MLR}}({\tilde{\bw}^{0}})}$ when compared to the state-of-the art algorithms.\\
\\
B \:\emph{Sparse Multinomial Logistic Regression}\\
a) In this simulation we fix the dimension of the feature vector $d$ to be $60$, number of samples $n$ to be $50$, number of classes $m$ to be $2$ and regularization parameter $\lambda$ as $0.25$ and compare the convergence speed of the proposed algorithm with the algorithms used to solve the Sparse MLR problem with $\ell_{1}$ regularization - MM based algorithm developed in \cite{mm}, ADMM algorithm \cite{admm} and the algorithm developed in \cite{scutari}. In the case of ADMM, $\tilde{\bw}$ was obtained by using BFGS algorithm, which was implemented using the available inbuilt function in Matlab. The elements of $\bx_{j}$ was randomly generated from a Standard Normal distribution with zero mean and unit variance. The algorithms were made to run till the condition in (\ref{condition}) was met with $f({\tilde{\bw}^{k}})$ defined as $l_{\textrm{MLR}}({\tilde{\bw}}) + \lambda\|\tilde{\bw}\|_{1}$. Fig. \ref{monotonic_sparse} shows the run time vs the objective value in log for the algorithms. 
From Fig. \ref{monotonic_sparse} it can be seen that the proposed algorithm takes lesser time to converge when compared to the other algorithms. We observed the same for different problem settings.\\
\\
b) In this simulation, we vary the size of $\tilde{\bw}$ and compare the performance of the proposed algorithm with the state-of-the art algorithms used to solve the Sparse MLR problem. The comparison is done based on how quickly the algorithms reduce the initial objective value to about $60\%$ of the initial objective value. The dimension $d$ was varied from $1000$ to $10000$ in steps of $1000$ and the number of samples $n$ and the number of classes $m$ was equal to $500$ and $2$, respectively. The elements of $\bx_{j}$ was randomly generated from a Standard Normal distribution with zero mean and unit variance. The initial value of $\tilde{\bw}$ was randomly generated from a uniform distribution from $[0,1]$ and was kept same for all the algorithms. Fig. \ref{vary_d_sparse} shows the performance of the algorithms for varying dimension $d$, number of samples $n$ equal to $500$, number of classses $m$ equal to $2$ and regularization parameter $\lambda$ equal to $0.5$.  From Fig. \ref{vary_d_sparse} it can be seen that the proposed algorithm takes the least time to converge when compared to the other algorithms. This could be because \textbf{PIANO} is the only algorithm which can parallely update each element of $\tilde{\bw}$ for the Sparse MLR problem.  \\
\\
c) In this simulation we fix the dimension of the feature vector $d$ equal to $12$, number of samples $n$ equal to $5$, number of classes $m$ equal to $2$ and show that the \textbf{PIANO} algorithm extended to solve the Sparse MLR problem with $\ell_{0}$ regularization is monotonic. Fig. \ref{l1_l0} shows the objective value vs time plot and as can be seen from the figure the proposed algorithm for the sparse MLR problem with $\ell_{0}$ regularization is monotonic.
\\
\\
C \:\emph{Real-life data}\\
In this subsection, we compare the performance of the algorithms with some benchmark data sets that are used to analyze classification algorithms. We briefly describe each data set here: 
\begin{enumerate}
\item{IRIS data set - It is a multivariate data set and contains fifty samples from each of the three species of the Iris flower - Iris Setosa, Iris Versicolour and Iris Virginica. Four features were measured from each sample - sepal length, sepal width, petal length and the petal width, all in centimeters. The task is to learn a classifier which could classify the type of species.  }
\item{Poker Hand data set - It is a multivariate data set containing $25010$ samples. Each sample is an example of a poker hand consisting of five playing cards drawn from a deck of $52$ cards. Each card is described using two features - suit and rank. Hence, the five cards are described by ten features. There is one class feature that describes the poker hand. The purpose of this data set is to learn a classifier which could distinguish the ten types of poker hands from each other. } 
\item{DB World emails data set - This data set contains $64$ emails from the DB World mailing list which announces conferences, jobs, books, software and grants. The  task is to train a classifier to learn to distinguish between announcement of conferences and everything else. Every email is represented as a vector containing $d$ binary values, where $d$ is taken to be $4702$ is the size of the vocabulary extracted from all the emails. The entry of the vector is one if the corresponding word belongs to the email and is zero otherwise. }
\item{URL Reputation data set - This data set contains $2.4$ million URLs which are collected from a large web mail provider over a period of $120$ days. The task is to train a classifier which could distinguish between malicious and benign websites. This data set contains $3.2$ million lexical and host-based features which were extracted from the URLs. We use a subset of this data to evaluate the performance of the algorithms i.e.,  we use $20000$ URLs and $50000$ features to compare the performance of the algorithms. } 
\end{enumerate}
The above data sets are available in \cite{data}. The summary of the statistics of the above data sets is given in Table. \ref{DS}. 
\begin{table}[h]
\centering
\caption {Summary of the Data Set Statistics}
\label{DS}
\begin{tabular}{|p{3cm}|p{1cm}|p{1.2cm}|p{1.4cm}|}
\hline
Data Set&Number of classes&Number of samples&Dimension of the feature vector\\
 \hline
IRIS&3&150&4\\
Poker Hand&10&25010&11\\
DB World emails&2&64&4702\\
URL Reputation&2&20000&50000\\
 \hline
\end{tabular} 
\end{table} 
The algorithms are compared based on how quickly the algorithms reduce the initial objective value to about $60\%$ of the initial objective value. When $n < d$ i.e. for DB World emails and URL Reputation data sets, we obtained the optimal $\tilde{\bw}$ by solving the Sparse MLR problem with $\ell_{1}$ regularization with $\lambda$ equal to $0.01$. For IRIS and Poker Hand data sets, since $n>d$, optimal $\tilde{\bw}$ was obtained by solving the MLR problem. The time taken by the algorithms is shown in Table. \ref{runtime}. Also, all the algorithms converged to the same $60\%$ of the initial objective value. 
\begin{table}[h]
\centering
\caption {Comparison of run time of the algorithms in seconds}
\label{runtime}
\begin{tabular}{|p{2.5cm}|p{1cm}|p{1cm}|p{0.5cm}|p{1cm}|p{1.3cm}|}
\hline
 Data Set&PIANO&MM&LC&ADMM&Facchinei\\
 \hline
IRIS&$0.03$&$0.06$&$0.04$&-&-\\
Poker Hand&$0.04$&$0.60$&$0.43$&-&-\\
DB World emails&$4.03\times10^{-4}$&$0.5$&-&$21.39$&$4.16$\\
URL Reputation&$0.076$&$116.8$&-&$>1$ hour&$16.16$\\
 \hline
\end{tabular} 
\end{table}
In the case of URL Reputation Data set, ADMM using BFGS algorithm ran into memory issues and hence $\tilde{\bw}$ was obtained using the memory efficient LBFGS algorithm. From Table. \ref{runtime} it can be seen that \textbf{PIANO} algorithm performs consistently well for all the data sets. 
\section{Conclusion}\label{sec:6}
In this paper, we proposed an iterative algorithm \textbf{PIANO} based on MM procedure to solve the Multinomial Logistic Regression problem. An attractive feature of \textbf{PIANO} is that it can parallely update each element of the weight vector $\tilde{\bw}$, which is useful when the number of features and classes are large. We then showed that \textbf{PIANO} can be easily extended to solve the Sparse Multinomial Logistic Regression problem with both $\ell_{0}$ and $\ell_{1}$ regularization. Computer simulations were conducted to compare the \textbf{PIANO} algorithm with the state-of-the art algorithms and was found that the proposed algorithm has faster speed of convergence.  
\bibliographystyle{IEEEtran} 
\bibliography{ref_reg}

\begin{thebibliography}{10}
\providecommand{\url}[1]{#1}
\csname url@samestyle\endcsname
\providecommand{\newblock}{\relax}
\providecommand{\bibinfo}[2]{#2}
\providecommand{\BIBentrySTDinterwordspacing}{\spaceskip=0pt\relax}
\providecommand{\BIBentryALTinterwordstretchfactor}{4}
\providecommand{\BIBentryALTinterwordspacing}{\spaceskip=\fontdimen2\font plus
\BIBentryALTinterwordstretchfactor\fontdimen3\font minus
  \fontdimen4\font\relax}
\providecommand{\BIBforeignlanguage}[2]{{%
\expandafter\ifx\csname l@#1\endcsname\relax
\typeout{** WARNING: IEEEtran.bst: No hyphenation pattern has been}%
\typeout{** loaded for the language `#1'. Using the pattern for}%
\typeout{** the default language instead.}%
\else
\language=\csname l@#1\endcsname
\fi
#2}}
\providecommand{\BIBdecl}{\relax}
\BIBdecl

\bibitem{bishop}
C.~M. Bishop, \emph{Pattern recognition and machine learning}.\hskip 1em plus
  0.5em minus 0.4em\relax springer, 2006.

\bibitem{book}
S.~Theodoridis, \emph{Machine learning: a Bayesian and optimization
  perspective}.\hskip 1em plus 0.5em minus 0.4em\relax Academic Press, 2015.

\bibitem{SVM}
V.~N. Vapnik, ``The nature of statistical learning,'' \emph{Theory}, 1995.

\bibitem{nn}
L.~K. Hansen and P.~Salamon, ``Neural network ensembles,'' \emph{IEEE
  transactions on pattern analysis and machine intelligence}, vol.~12, no.~10,
  pp. 993--1001, 1990.

\bibitem{hmm}
P.~R. Runkle, P.~K. Bharadwaj, L.~Couchman, and L.~Carin, ``Hidden markov
  models for multiaspect target classification,'' \emph{IEEE Transactions on
  Signal Processing}, vol.~47, no.~7, pp. 2035--2040, 1999.

\bibitem{bayes}
I.~Rish \emph{et~al.}, ``An empirical study of the naive bayes classifier,'' in
  \emph{IJCAI 2001 workshop on empirical methods in artificial intelligence},
  vol.~3, no.~22, 2001, pp. 41--46.

\bibitem{hyperspectral}
J.~Li, J.~M. Bioucas-Dias, and A.~Plaza, ``Semisupervised hyperspectral image
  segmentation using multinomial logistic regression with active learning,''
  \emph{IEEE Transactions on Geoscience and Remote Sensing}, vol.~48, no.~11,
  pp. 4085--4098, 2010.

\bibitem{semisupervised}
------, ``Semisupervised hyperspectral image classification using soft sparse
  multinomial logistic regression,'' \emph{IEEE Geoscience and Remote Sensing
  Letters}, vol.~10, no.~2, pp. 318--322, 2012.

\bibitem{text1}
A.~Genkin, D.~D. Lewis, and D.~Madigan, ``Large-scale bayesian logistic
  regression for text categorization,'' \emph{Technometrics}, vol.~49, no.~3,
  pp. 291--304, 2007.

\bibitem{text2}
G.~Ifrim, G.~Bakir, and G.~Weikum, ``Fast logistic regression for text
  categorization with variable-length n-grams,'' in \emph{Proceedings of the
  14th ACM SIGKDD international conference on Knowledge discovery and data
  mining}.\hskip 1em plus 0.5em minus 0.4em\relax ACM, 2008, pp. 354--362.

\bibitem{biomedical1}
I.~Kurt, M.~Ture, and A.~T. Kurum, ``Comparing performances of logistic
  regression, classification and regression tree, and neural networks for
  predicting coronary artery disease,'' \emph{Expert systems with
  applications}, vol.~34, no.~1, pp. 366--374, 2008.

\bibitem{biomedical2}
G.~C. Cawley and N.~L. Talbot, ``Gene selection in cancer classification using
  sparse logistic regression with bayesian regularization,''
  \emph{Bioinformatics}, vol.~22, no.~19, pp. 2348--2355, 2006.

\bibitem{data}
\BIBentryALTinterwordspacing
D.~Dua and C.~Graff, ``{UCI} machine learning repository,'' 2017. [Online].
  Available: \url{http://archive.ics.uci.edu/ml}
\BIBentrySTDinterwordspacing

\bibitem{large_data_1}
I.~Partalas, A.~Kosmopoulos, N.~Baskiotis, T.~Arti{\`{e}}res, G.~Paliouras,
  {\'{E}}.~Gaussier, I.~Androutsopoulos, M.~Amini, and P.~Gallinari, ``{LSHTC:}
  {A} benchmark for large-scale text classification,'' \emph{CoRR}, vol.
  abs/1503.08581, 2015.

\bibitem{large_data_2}
J.~Deng, W.~Dong, R.~Socher, L.-J. Li, K.~Li, and L.~Fei-Fei, ``Imagenet: A
  large-scale hierarchical image database,'' in \emph{2009 IEEE conference on
  computer vision and pattern recognition}.\hskip 1em plus 0.5em minus
  0.4em\relax Ieee, 2009, pp. 248--255.

\bibitem{vapnik}
V.~Vapnik, \emph{The nature of statistical learning theory}.\hskip 1em plus
  0.5em minus 0.4em\relax Springer science \& business media, 2013.

\bibitem{IRLS}
D.~P. O’Leary, ``Robust regression computation using iteratively reweighted
  least squares,'' \emph{SIAM Journal on Matrix Analysis and Applications},
  vol.~11, no.~3, pp. 466--480, 1990.

\bibitem{mm}
B.~Krishnapuram, L.~Carin, M.~A. Figueiredo, and A.~J. Hartemink, ``Sparse
  multinomial logistic regression: Fast algorithms and generalization bounds,''
  \emph{IEEE transactions on pattern analysis and machine intelligence},
  vol.~27, no.~6, pp. 957--968, 2005.

\bibitem{mm_parallel}
S.~Gopal and Y.~Yang, ``Distributed training of large-scale logistic models,''
  in \emph{International Conference on Machine Learning}, 2013, pp. 289--297.

\bibitem{admm}
S.~Boyd, N.~Parikh, E.~Chu, B.~Peleato, J.~Eckstein \emph{et~al.},
  ``Distributed optimization and statistical learning via the alternating
  direction method of multipliers,'' \emph{Foundations and
  Trends{\textregistered} in Machine learning}, vol.~3, no.~1, pp. 1--122,
  2011.

\bibitem{scutari}
F.~Facchinei, G.~Scutari, and S.~Sagratella, ``Parallel selective algorithms
  for nonconvex big data optimization,'' \emph{IEEE Transactions on Signal
  Processing}, vol.~63, no.~7, pp. 1874--1889, 2015.

\bibitem{prob}
A.~Papoulis and S.~U. Pillai, \emph{Probability, random variables, and
  stochastic processes}.\hskip 1em plus 0.5em minus 0.4em\relax Tata
  McGraw-Hill Education, 2002.

\bibitem{feature_selection}
A.~Y. Ng, ``Feature selection, l 1 vs. l 2 regularization, and rotational
  invariance,'' in \emph{Proceedings of the twenty-first international
  conference on Machine learning}.\hskip 1em plus 0.5em minus 0.4em\relax ACM,
  2004, p.~78.

\bibitem{l0_approx_1}
H.~A. Le~Thi, H.~M. Le, T.~P. Dinh \emph{et~al.}, ``A dc programming approach
  for feature selection in support vector machines learning,'' \emph{Advances
  in Data Analysis and Classification}, vol.~2, no.~3, pp. 259--278, 2008.

\bibitem{l0_approx_2}
H.~M. Le, H.~A. Le~Thi, and M.~C. Nguyen, ``Sparse semi-supervised support
  vector machines by dc programming and dca,'' \emph{Neurocomputing}, vol. 153,
  pp. 62--76, 2015.

\bibitem{sir}
Y.~Sun, P.~Babu, and D.~P. Palomar, ``Majorization-minimization algorithms in
  signal processing, communications, and machine learning,'' \emph{IEEE
  Transactions on Signal Processing}, vol.~65, no.~3, pp. 794--816, 2016.

\bibitem{tutorial}
D.~R. Hunter and K.~Lange, ``A tutorial on {MM} algorithms,'' \emph{The
  American Statistician}, vol.~58, no.~1, pp. 30--37, 2004.

\bibitem{boyd}
S.~Boyd and L.~Vandenberghe, \emph{Convex optimization}.\hskip 1em plus 0.5em
  minus 0.4em\relax Cambridge university press, 2004.

\bibitem{jensen1}
J.~L. W.~V. Jensen, ``Om konvekse funktioner og uligheder imellem
  middelvaerdier,'' \emph{Nyt tidsskrift for matematik}, vol.~16, pp. 49--68,
  1905.

\bibitem{jensen2}
J.~L. W.~V. Jensen \emph{et~al.}, ``Sur les fonctions convexes et les
  in{\'e}galit{\'e}s entre les valeurs moyennes,'' \emph{Acta mathematica},
  vol.~30, pp. 175--193, 1906.

\bibitem{convergence}
M.~Razaviyayn, M.~Hong, and Z.-Q. Luo, ``A unified convergence analysis of
  block successive minimization methods for nonsmooth optimization,''
  \emph{SIAM Journal on Optimization}, vol.~23, no.~2, pp. 1126--1153, 2013.

\end{thebibliography}
\end{document}